%% file: main.tex
\documentclass[10pt,twocolumn,letterpaper]{article}

\usepackage{cvpr}              % To produce the CAMERA-READY version
\usepackage[accsupp]{axessibility}

\input{preamble}
\usepackage{lipsum}
\usepackage{footnote}
\usepackage{subcaption}

\makeatletter
\def\blfootnote{\gdef\@thefnmark{}\@footnotetext}
\makeatother
\def\paperID{2505} % *** Enter the Paper ID here
\def\confName{CVPR}
\def\confYear{2024}

\title{Visual Delta Generator with Large Multi-modal Models \\ for Semi-supervised Composed Image Retrieval}

\author{Young Kyun Jang\thanks{Authors contributed equally.} \textsuperscript{1}, Donghyun Kim\footnotemark[1] \textsuperscript{2}, Zihang Meng\textsuperscript{1}, Dat Huynh\textsuperscript{1}, and Ser-Nam Lim\textsuperscript{3} \\
Meta AI\textsuperscript{1}, Korea University\textsuperscript{2}, University of Central Florida\textsuperscript{3}\\
}

\begin{document}
\maketitle
\input{sec/0_abstract}    
\input{sec/1_intro}
\input{sec/2_related_work}
\input{sec/3_method}

\input{sec/4_experiment}

\input{sec/5_conclusion}

{
    \small
    \bibliographystyle{ieeenat_fullname}
    \bibliography{main}
}

\input{sec/X_suppl}

\end{document}

%% file: preamble.tex
\usepackage[dvipsnames]{xcolor}

\definecolor{cvprblue}{rgb}{0.21,0.49,0.74}
\usepackage[pagebackref,breaklinks,colorlinks,citecolor=cvprblue]{hyperref}
\usepackage{tcolorbox}
\usepackage{caption}
\usepackage{multirow}
\usepackage{booktabs}
\usepackage{adjustbox}
\usepackage{xcolor}
\usepackage{colortbl}
\usepackage{makecell}

\definecolor{grey}{rgb}{0.9, 0.9, 0.9}
\definecolor{mgt}{rgb}{0.8, 0.1, 0.8}

\definecolor{blu}{rgb}{0.1, 0.8, 0.1}

\usepackage{soul}

\definecolor{supervisedBase}{rgb}{0.1, 0.1, 0.8} 
\definecolor{unsupervisedBase}{rgb}{0.8, 0.1, 0.1}
\definecolor{oursColorBase}{rgb}{0.9,0.9,0.9}

\colorlet{supervisedColor}{supervisedBase!10!white}
\colorlet{unsupervisedColor}{unsupervisedBase!10!white}
\colorlet{oursColor}{oursColorBase!50!white}

\definecolor{azure(colorwheel)}{rgb}{0.0, 0.5, 1.0}

\definecolor{codegreen}{rgb}{0,0.6,0}
\definecolor{codegray}{rgb}{0.5,0.5,0.5}
\definecolor{codepurple}{rgb}{0.58,0,0.82}
\definecolor{backcolour}{rgb}{0.95,0.95,0.92}

\usepackage{algorithm}
\usepackage{algorithmic}
\usepackage{listings}

\lstdefinestyle{mystyle}{
    commentstyle=\color{codegreen},
    keywordstyle=\color{magenta},
    numberstyle=\tiny\color{codegray},
    stringstyle=\color{codepurple},
    basicstyle=\ttfamily\footnotesize,
    breakatwhitespace=false,         
    breaklines=true,                 
    captionpos=b,                    
    keepspaces=true,                 
    numbers=left,                    
    numbersep=5pt,                  
    showspaces=false,                
    showstringspaces=false,
    showtabs=false,                  
    tabsize=2
}

%% file: sec/0_abstract.tex
\begin{abstract}
Composed Image Retrieval (CIR) is a task that retrieves images similar to a query, based on a provided textual modification. Current techniques rely on supervised learning for CIR models using labeled triplets of the \textless reference image, text, target image\textgreater. These specific triplets are not as commonly available as simple image-text pairs, limiting the widespread use of CIR and its scalability. On the other hand, zero-shot CIR can be relatively easily trained with image-caption pairs without considering the image-to-image relation, but this approach tends to yield lower accuracy. We propose a new semi-supervised CIR approach where we search for a reference and its related target images in auxiliary data and learn our large language model-based Visual Delta Generator (VDG) to generate text describing the visual difference (\ie, visual delta) between the two. VDG, equipped with fluent language knowledge and being model agnostic, can generate pseudo triplets to boost the performance of CIR models. Our approach significantly improves the existing supervised learning approaches and achieves state-of-the-art results on the CIR benchmarks.

 % In addition, we propose a new learning objective that could align the target and reference \dktodo{additional loss}.

\end{abstract}

%% file: sec/1_intro.tex
\section{Introduction}
\label{sec:intro}
% \dktodo{LLM or The LLM?}

Image-to-image or text-to-image retrieval, where a query image/text is used to retrieve similar ones from a gallery, has grown into a pivotal research field with many practical applications \cite{IR_survey}. However, relying solely on image queries is limiting, as they primarily retrieve similar images, making it challenging to understand the user's intent for modifications in the results. On the other hand, relying solely on text queries can also be restrictive, as it may not effectively convey the user's desired detailed visual contents. To address this, Composed Image Retrieval (CIR) was introduced \cite{CIRR,fashionIQ,Combiner,Pic2word}. CIR seeks to retrieve images using a query that combines both an image and a textual description of the user's intent (referred to as the \textit{visual delta}), which allows more flexible retrieval. Due to the convenience and diverse applicability of CIR, it has attracted increased attention recently for a variety of real-world applications.

% especially for a variety of real-world applications such as web search or recommendation systems.

\begin{figure}[!t]
\centering
  \subcaptionbox{CIR triplet generation with human supervision (expensive).
  \label{fig:intro_a}}{\includegraphics[height=4cm]{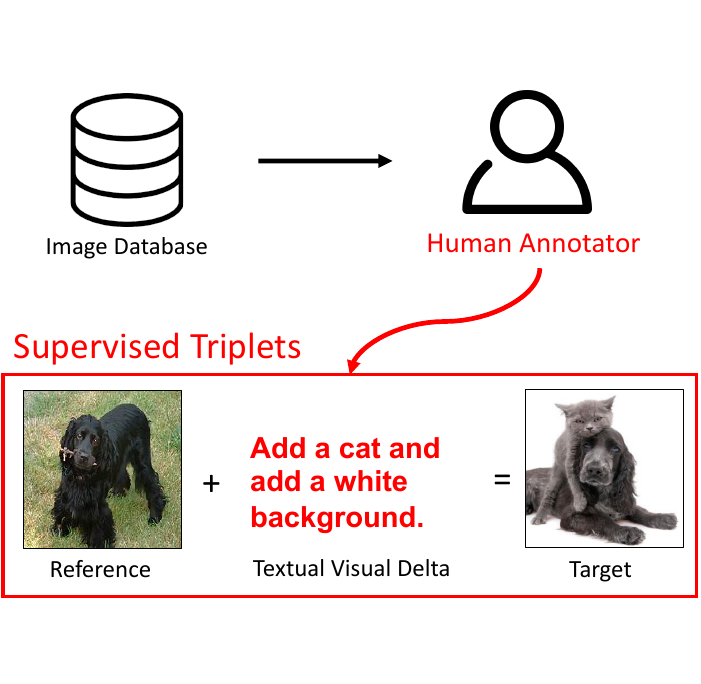}}
  \subcaptionbox{Visual Delta Generator for generating pseudo triplets.
  \label{fig:intro_b}}{\includegraphics[height=4cm]{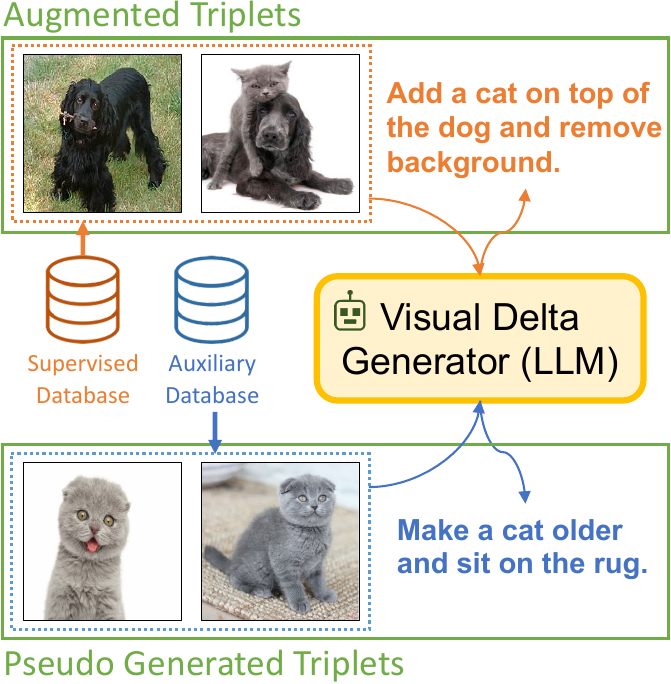}}
\caption{An illustration of the data preparation process of (a) conventional supervised Composed Image Retrieval (CIR) \vs (b) our proposed semi-supervised CIR. While supervised CIR struggles to scale up due to high annotation costs, our semi-supervised method offers a cost-effective and scalable solution. It augments training samples efficiently by generating pseudo triplets through our Large Language Model (LLM)-based Visual Delta Generator.}
\label{fig:intro}
\end{figure}

% \begin{figure}[!t]
% \centering
% \includegraphics[width=0.9\linewidth]{intro.png}
% \caption{{\color{red} Zihang: should the arrow of "1. select pairs" be the opposite direction? and the minus sign feels a little confusing}Illustration of (a) conventional supervised approach for Composed Image Retrieval (CIR) \vs (b) our proposed semi-supervised approach for CIR. Most of the approaches for CIR were proposed under the supervised learning setup, which is hard to scale due to the expensive annotation cost. We propose a semi-supervised learning approach for CIR by generating pseudo triplets with our Visual Delta Generator (VDG).}
% \label{fig:intro}
% \end{figure}

Existing research on CIR has been developed under two major settings: (1) Supervised CIR: Learning with supervised triplets (\ie \textit{\textless reference image, visual delta, target image \textgreater}) \cite{TIRG,CIRPLANT,Combiner,ARTEMIS,CASE} as shown in Fig. \ref{fig:intro} (a), and (2) Zero-shot CIR: Learning with massive noisy \textit{\textless image, textual caption \textgreater} pairs \cite{Pic2word,SEARLE,CoVR}, without any CIR supervision. %Obviously, 
Supervised CIR would obviously yield much higher accuracy in retrieval %and allows a model to learn relative representation from the reference image and relative visual deltas to retrieve the target image. However, it 
but requires expensive two-stage data collection processes - collecting pairs of related reference and target images, and then annotating them with visual delta that 
depicts the difference between them.
%modifies the reference to the target. 
On the other hand, zero-shot CIR does not incur additional labeling costs and utilizes web-collected noisy image text caption pairs directly. However, it has a much lower performance bar compared to supervised approaches and lacks the ability to specialize in specific CIR domain tasks.

%A model trained on tons of noisy image-caption pairs can be used for zero-shot CIR inference.

\begin{figure}[!t]
\centering
\includegraphics[width=0.9\linewidth]{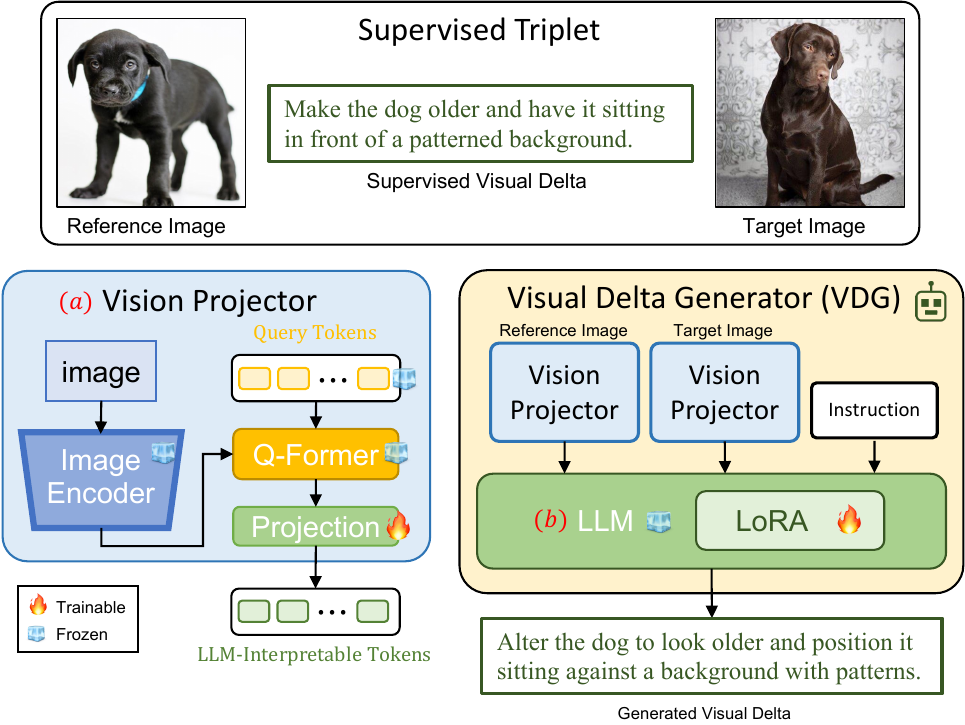}
\caption{An overview of the VDG tuning process. It includes \textcolor{red}{(a)} a vision projector and \textcolor{red}{(b)} a Large Language Model (LLM). The VDG is trained to produce visual delta that accurately describes the difference between a reference image and its corresponding target image.
% Instructional tuning pipeline of Visual Delta Generator.
}
\label{fig:VDG_training}
\end{figure}

In this paper, we investigate a class of CIR called \textit{semi-supervised CIR}, blending supervised and unsupervised samples to enhance generalization (Fig. \ref{fig:intro} (b)). This method focuses on boosting CIR performance in specific retrieval domains by %augmenting visual delta and 
creating new triplets from unsupervised data. Building on this concept, we introduce a novel technique, Visual Delta Generator (VDG), designed to tap into the extensive natural language capabilities of Large Language Models (LLMs) \cite{bert,gpt-3,llama,llama2}. Our approach involves projecting reference and target images from supervised CIR triplets into the language embedding space, making them suitable inputs for the LLM. We then fine-tune the model by using prompts such as \textit{`Describe the differences between images.'} to induce it to yield human-like %textual description, i.e., the 
visual delta as illustrated in Fig. \ref{fig:VDG_training}. Furthermore, we employ a parameter-efficient fine-tuning technique, LoRA \cite{LoRA}, on the LLM. This choice of design effectively enhances the quality of visual deltas while also preserving the LLM's original capabilities, without harming its inherent knowledge.

After the VDG is trained, %the LLM 
it knows how to distinguish between a given reference and target image and produce visual delta as a textual response. If we forward two similar images with different compositions, we can thus obtain the corresponding visual delta easily with the VDG. %LLM. To facilitate this, 
This allows us to achieve two purposes. First, we can now augment existing CIR triplets by adding generated visual deltas to pairs of reference and target images from the training set. Second, we can also harvest new reference-target pairs from an unlabeled database based on visual similarity, %. Then, 
after which we forward these images to VDG to configure new pseudo triplets for %semi-supervised 
CIR training. Note that our VDG is model agnostic -- it simply increases the number of triplet candidates for training any given supervised CIR baselines. This strategy strikes a balance between maintaining the integrity of supervisory concepts derived from a supervised dataset and the capacity for effortless expansion using new, unlabeled image samples. It's a cost-effective and scalable solution, ensuring uniformity in annotations across extensive datasets. By generating pseudo triplets with VDG, we significantly reduce annotation costs and enhance the performance of CIR models trained solely on supervised learning, as well as those trained without supervised triplets. Our approach leads to state-of-the-art results in CIR benchmarks.

% The approach leverages the versatility of LLMs, establishing a sturdy and adaptable annotation process that could significantly improve CIR practices. 

% Furthermore, in order not to change the visual embeddings for retrieval to avoid re-constructing gallery with CIR fine-tuned model and perform retrieval on existing gallery, we configure our framework with a frozen Vision Transformer (ViT) and a trainable composer based on image-grounded text encoder of vision-language pretraining model BLIP, to produce composed embedding for retrieval. We adopt basic contrastive loss between composed embedding generated by reference image and visual delta and vision embedding from target image. Additionally, we propose image-visual delta matching loss on the key idea that visual delta should contain important context that reveals in target image, therefore, we make composer to learn fine-grained alignment between target image patch tokens and visual delta.

The key advantages of our semi-supervised CIR include:

\begin{itemize}
\item To the best of our knowledge, we are the first to transfer knowledge from Large Language Models (LLMs) and connect it with semi-supervised Composed Image Retrieval (CIR).
\item We propose a novel Visual Delta Generator (VDG) that generates synthetic visual deltas for augmenting the supervised dataset and allowing the integration of an auxiliary image gallery for CIR model training.
\item Comprehensive experimental results confirm the effectiveness of our method, demonstrating state-of-the-art retrieval rankings and showcasing the potential of our work.
\end{itemize}

%% file: sec/2_related_work.tex
\section{Related Work}
\label{sec:Related Work}

\paragraph{Composed Image Retrieval.}
The field of image retrieval has captured the interest of many researchers in our community \cite{IR_survey,UNITER}. One notable area that has seen much progress recently is Composed Image Retrieval (CIR), a problem that focuses on retrieving images that best match a given pair of a query image and textual intent.
%CIR is unique because, unlike traditional content-based or text-based uni-modal searches, it requires an understanding of both image and text modalities.  Additionally, it differs from tasks of single image compositional learning like image captioning \cite{image_captioning} or VQA \cite{VQA}, as it necessitates linking the relationship between two different images using text representations.
Supervised CIR methods \cite{TIRG, ARTEMIS, CIRPLANT, Combiner} are trained on human-annotated triplets, consisting of a reference image, a target image, and their textual difference. On the other hand, zero-shot CIR \cite{Pic2word, SEARLE, PALAVRA, gu2023compodiff} operates without relying on human-guided descriptions of the differences between the two images. Instead, it uses noisy image-text pairs, aiming to find a function that can translate images into words. This approach, designed to discriminate subtle differences between images based solely on text captions, poses challenges, but also scales well, making it easy to add more data for CIR training.  Addressing the limitations of both approaches, we explore the field of semi-supervised learning-based retrieval in this work, leveraging the strengths of both labeled and unlabeled data to enhance retrieval performance.

\paragraph{Semi-supervised Learning.}
Semi-supervised learning has been an active research topic in visual recognition for a long time~\cite{GPQ,wen2023self,decouplednet,sohn2020fixmatch,learning2006semi,lee2013pseudo,berthelot2019mixmatch,laine2016temporal,miyato2018virtual,tarvainen2017mean}. Semi-supervised learning can be roughly categorized into two groups, consistency learning~\cite{laine2016temporal,miyato2018virtual,tarvainen2017mean} and pseudo-labeling based learning~\cite{lee2013pseudo,sohn2020fixmatch}. Consistency-based methods, as the name implies, encourage consistency in the output of the model by adding noise to model weights or using Exponential Moving Average (EMA) from a teacher model to a student model. In pseudo-labeling based methods, hard/soft pseudo-labels obtained from a pretrained model are assigned to unlabeled images~\cite{grandvalet2004semi,sohn2020fixmatch}. These pseudo-labels can be filtered using confidence thresholding or multi-view consistency. However, neither consistency-based nor pseudo labels methods are directly applicable to CIR. This is because the relative textual descriptions of the corresponding visual differences (visual delta) needed in CIR are not the intended outputs of these methods. In this work, we propose the Visual Delta Generator (VDG), a multi-modal pseudo-label generator. VDG processes two input images and generates text that describes their visual differences, making it an ideal candidate for constructing pseudo triplets for CIR.

\paragraph{Multi-modal Models for Image-Text Retrieval.} 
Models like CLIP \cite{CLIP} and BLIP \cite{BLIP} have showcased the advantages of training models on extensive image-text pairs, enabling precise alignment between language and vision representations that is crucial for image-text retrieval. Building on this, there have been significant advancements in utilizing Large Language Models (LLMs) for enhanced vision-language understanding \cite{LLM_survey, BLIP-2, InstructBLIP, Flamingo, llava}. Especially for CIR, Fromage \cite{Fromage} utilizes LLMs to directly produce embeddings for retrieval, allowing cross-modal compositional search with a single image and textual intent. CoVR \cite{CoVR} and SEARLE \cite{SEARLE} apply LLMs to generate visual deltas using image captions without incorporating visual data, which limits the generation of accurate CIR training samples. Our proposed method overcomes these challenges by fine-tuning LLMs with the integration of a pretrained vision-language alignment module. This integration empowers LLMs to perceive and comprehend images, making them applicable even in scenarios with only image datasets. The method excels in generating accurate visual deltas, a key factor in training efficient external CIR models. These enhancements optimize the use of LLMs while ensuring computational efficiency, thereby expanding the versatility of LLMs in image-related tasks.

%% file: sec/3_method.tex
\section{Method}
\label{sec:Method}

\paragraph{Overview.} Our goal is to establish a semi-supervised Composed Image Retrieval (CIR) system that merges image reference features with user textual descriptions to retrieve images from a large-scale database. We face a challenge in the limited availability of supervised triplets necessary for robust CIR model training. To overcome this limitation, we introduce a novel semi-supervised approach for CIR by leveraging an instruction-tuned Large Language Model (LLM), which we call Visual Delta Generator (VDG). The VDG learns to discriminate differences between two images and produces a textual response. This capability allows us to generate additional CIR training triplets, which in turn contributes to the development of more robust CIR models. Sec.~\ref{sec:VDG_training} provides detailed insights into the construction of the VDG. Sec.~\ref{sec:pseudo_triplet_generation} describes the pseudo triplet generation process. Training of CIR models with pseudo triplets and our additional objective function for better optimization are described in Sec.~\ref{sec:cir_training}.

\subsection{Visual Delta Generator Training}
\label{sec:VDG_training}

While semi-supervised learning methods have been actively developed for standard visual recognition tasks, these cannot be directly applied to CIR. In CIR, pseudo-label generation requires a detailed semantic understanding of two separate images 
%which in result 
such that their difference can be automatically expressed in the form of text.
%should be represented as a form of text
We leverage vision-language pretraining models and LLMs to achieve the requirements. With the huge success of LLMs, there are approaches that aim to utilize their understanding of the language domain for improving vision tasks. Particularly, LLaVA \cite{llava} and InstructBLIP \cite{InstructBLIP} which are trained on top of the chat-bot style instruction tuned LLM, Vicuna \cite{Vicuna}, have shown interesting results on vision-language tasks. Inspired by these, we propose VDG, which allows the LLM to take two images (reference, target) with similar contents and discriminate their difference in the form of text response (visual delta) as shown in Fig.~\ref{fig:VDG_training}.  

% Each sample is configured as triplet: two images and their corresponding visual delta, and we use it specifically for the instructional fine-tuning of the LLM.
% For this purpose, we employ an existing supervised dataset comprised of CIR training samples.

First, to enable the LLM to interpret images, we use the Querying Transformer (Q-Former) motivated by InstructBLIP \cite{InstructBLIP} to prepare images for the LLM input (\ie, \textit{Vision Projector} (VP) in Fig.~\ref{fig:VDG_training}\textcolor{red}{(a)}). Q-Former, a transformer encoder \cite{bert,vit}, processes a fixed set of 32 learnable query tokens (embeddings). These tokens are modified through self-attention layers and interact with image features of Vision Transformer (ViT)~\cite{vit} through cross-attention layers. As a result, the query tokens are infused with the visual information from the provided image, making them suitable for LLM processing. 

% To maintain the flexibility of our system in accommodating future advancements and iterations in Large Language Models (LLMs), we chose LLaMA2 for our foundational LLM, diverging from the original vicuna-13B model utilized in InstructBLIP. This strategic choice anticipates the continuous evolution of LLMs, ensuring our methodology stays relevant and adaptable. However, integrating the Q-Former with a new LLM poses compatibility challenges, as Q-Former was tailored for specific models.

\paragraph{Stage 1: Alignment.} 
%Direct use of the outputs of the Vision Projector (VP) have discrepancy with tokenized word embeddings of LLM inputs if we utilize new LLM with different vocabulary, which we choose LLaMA2 \cite{llama2} for our foundational LLM for text generation. 
The outputs of the VP are inherently not aligned with the tokenized word embeddings of the inputs to the LLM, which we chose to be LLaMA2 \cite{llama2} in this work. Alignment between the VP and LLM needs to be attained by fine-tuning trainable projection of VP in Fig.~\ref{fig:VDG_training}\textcolor{red}{(a)} as was done in LLaVA ~\cite{llava}. Specifically, we employ large-scale image-caption pairs to foster alignment between the image representations understood by the VP and the LLM. This step includes minimizing standard next token prediction loss which is generally used to train the decoder-based LLM as:

\begin{equation}
L_{LLM} = -\sum_{t=1}^{T} \log P(\mathrm{w}_t |x,I_{inst},\mathrm{w}_{(1,..,t\text{-}1)}; \theta_{proj}),
\label{eqn:LLM_loss}
\end{equation}

\noindent where $\mathrm{w}_{t}$ denotes a token at time step $t$, $T$ is the length of the sequence, $P(\cdot)$ is the probability assigned by the model to the actual token $\mathrm{w}_t$ given with the image $x$, instruction $I_{inst}$ and previous tokens $\mathrm{w}_{(1,..,t\text{-}1)}$, and $\theta_{proj}$ is parameters of projection layer. Following standard practice, the prediction for each token is computed using a softmax over the vocabulary of the LLM, and the cross-entropy loss is computed between the predicted probabilities and the textual token labels as a classification.

\paragraph{Stage 2: Instruction Tuning.} In this stage, we conduct instruction tuning to equip our LLM with the ability to understand image pairs and produce visual delta, as shown in Fig.~\ref{fig:VDG_training}\textcolor{red}{(b)}. Referring to Fig.~\ref{fig:instruction_tuning}, we train the LLM with an instruction (\ie, ``\textbf{Request}'') to generate the visual delta in textual format (\ie, ``\textbf{Response}''). We forward two distinct images (\ie, ``\textbf{Reference}'' and ``\textbf{Target}'') into the LLM via the VP. We employ a parameter-efficient fine-tuning technique, LoRA \cite{LoRA}, directly on the LLM. This fine-tuning process is designed to provide ``extra room'' for the LLM to undertake new tasks, all without compromising its foundational capabilities. In this stage, we introduce a specific prompt structure to the LLM as below: 
% \dk{train Project-layer + Lora in LLM}

\begin{figure}[h]
\centering
\scalebox{0.9}{ % Adjust the scale value as needed
\begin{tcolorbox}[width=0.50\textwidth, colback=gray!10!white, colframe=gray!75!black, left=1mm, right=1mm, top=1mm, bottom=1mm]
\small % Makes the text size smaller; you can also use \scriptsize or \tiny for even smaller text
\textbf{Request:} Analyze given reference and target images and provide a description that transforms the reference to match the target. \\
\textbf{Reference:} \textless Reference image query tokens\textgreater \\
\textbf{Target:} \textless Target image query tokens\textgreater \\
\textbf{Response: \textless\textcolor{red}{Visual delta (text label)}}\textgreater
\end{tcolorbox}
}
\vspace{-2mm}
\caption{A template prompt for VDG instruction tuning.}
\label{fig:instruction_tuning}
\end{figure}

\noindent which guides the LLM to generate the corresponding visual delta. We leverage the same training loss from Eqn. \ref{eqn:LLM_loss} to train LoRA parameters.

% \dktodo{Need to rewrite since model parameters changed}. 

\subsection{Pseudo Triplet Generation for CIR}
\label{sec:pseudo_triplet_generation}
After training VDG, we can generate visual delta of two images, which can be used to form pseudo triplets for CIR. However, it is important to choose two images that not only share certain attributes and similarities but also present other distinct attributes (\ie, visual delta). To gather suitable pairs of reference and target images, we utilize an image encoder to select them from an auxiliary image gallery, which we denote as $G'$. Note that, we notate \textit{upper strophe} ($'$) on samples and embeddings that are obtained from $G'$, in the following.

\begin{figure}[!t]
\centering
\includegraphics[width=0.9\linewidth]{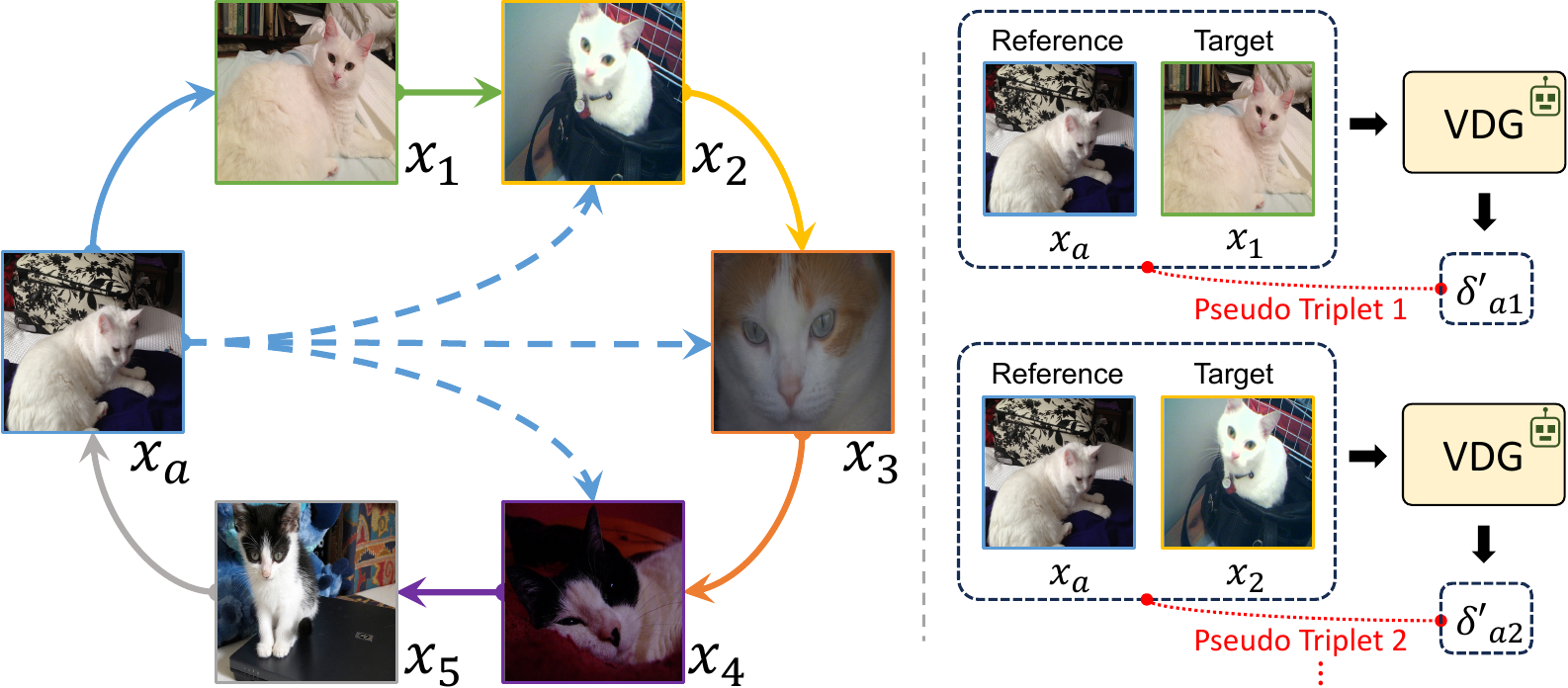}
\caption{The process of pseudo triplet generation. First, an image subgroup is constructed based on visual similarity (left). Then, paired reference and target images are fed into the VDG to generate the visual delta, completing the triplet formation (right).}
\label{fig:Pseudo_triplet_generation}
\end{figure}

Following the strategy in CIRR \cite{CIRR}, we start with an anchor image $x_{a}$ and retrieve the top 20 images from $G'$ using cosine similarity scores between ResNet 152 \cite{ResNet} embeddings, pretrained on ImageNet \cite{ImageNet}. We exclude images with scores above 0.94 and sequentially add images to a subgroup of size 6, skipping those within a 0.002 score of the previously included image. As depicted in Fig. \ref{fig:Pseudo_triplet_generation}, we establish dense connections between all pairs (both consecutive, represented by the outer circle, and non-consecutive, represented by the dotted inner connections), while ensuring no overlaps. The arrow's starting point denotes the \textit{reference}, while its endpoint indicates the \textit{target}. %Both connections between consecutive (represented by the outer circle) and non-consecutive pairs (shown by dotted inner connections) are considered. 
Given images $x'^r_i$ and $x'^t_j$, VDG produces $\delta'_{i,j}$. This allows us to formulate the pseudo triplet as $\{x'^r_i, x'^t_j, \delta'_{i,j}\}$.

\subsection{Semi-supervised CIR Training}

\paragraph{Preliminaries.} Suppose we have access to a CIR dataset of triplets: $D=\{(x^r_i, x^t_j, \delta_{n(i,j)})\}_{n=1}^N$ where $x^r_i$, $x^t_j$ denote reference and target images, respectively, $\delta_{n(i,j)}$ represents their visual delta, and $N$ denotes the total triplet counts.

% Note that a given image can be paired with multiple other target images with different $\delta$ and form a triplet, thus the total count of unique $x^r$ and $x^t$ images in dataset $D$ is fewer than the triplet count $N$.

In the pursuit of enhancing CIR through a semi-supervised approach, our method's strength lies in its model-agnosticism, allowing for seamless integration with a variety of CIR models. Following recent trends, we opt to use the encoders from vision-language pretraining models~\cite{CLIP, InstructBLIP, BLIP, BLIP-2}, notably CLIP and BLIP, as our baseline backbones. These encoders are naturally equipped to understand and convert both visual and textual elements into a joint embedding space, making them suitable for CIR tasks.

The image encoder takes patchified image token embeddings $x=[\mathrm{x}_1, ... \mathrm{x}_{K_{img}}]$ with the learnable image \textit{cls} token embedding $[\mathrm{x}_{cls}]$, and outputs the visual feature embeddings $E_{img}(\mathrm{x}_{cls},x)=[\mathrm{\hat{x}}_{cls}, \mathrm{\hat{x}}_1, ..., \mathrm{\hat{x}}_{K_{img}}]$ where $\mathrm{\hat{x}}\in R^{d_i}$ of ${d_i}$ dimension, and $K_{img}$ denotes the number of generated tokens from each image. The text encoder processes tokenizer output embedding of visual delta $\delta=[\mathrm{z}_1, ... \mathrm{z}_{K_{txt}}]$ with the learnable text \textit{cls} token embedding $[\mathrm{z}_{cls}]$ to produce its textual feature embeddings $E_{txt}(\mathrm{z}_{cls},\delta)=[\mathrm{\hat{z}}_{cls}, \mathrm{\hat{z}}_1, ..., \mathrm{\hat{z}}_{K_{txt}}]$, where $\mathrm{\hat{z}}\in R^{d_t}$ of ${d_t}$ dimension, and $K_{txt}$ denotes the number of generated text tokens from each visual delta.

% Per standard practice, we employ $\mathrm{\hat{x}}_{cls}$ and $\mathrm{\hat{w}}_{cls}$ to serve as the representative embeddings for the image and text samples, respectively.

\paragraph{Model Architecture.} To carry out CIR, we establish a fusion function $f$ that takes a reference image $x^r$ and a visual delta $\delta$, producing a composed embedding $\mathbf{c}$ as: $f(x^r, \delta)=\mathbf{c}$, and notates its trainable components as $f_{\theta}$. We utilize two well-known backbones, CLIP and BLIP, to configure $f$. In the case of CLIP, we employ the text encoder $E_{txt:\theta}$ and an additional Combiner module \cite{Combiner} $\mathcal{C}_{\theta}(\mathrm{\hat{x}}^r_{cls}, \mathrm{\hat{z}}_{cls})$ that outputs $\mathbf{c}$ to be the components of $f_{\theta}$ (\ie, $f^{\text{CLIP}}_{\theta}=\{E_{txt:\theta}, \mathcal{C}_{\theta}\}$). The Combiner is designed to optimally blend $\mathrm{\hat{x}}_{cls}$ and $\mathrm{\hat{z}}_{cls}$ carefully weighing their individual impacts while adeptly mixing them. In the BLIP case, we exclusively use BLIP's text encoder, grounded in the image, and designate it as the trainable $E_{txt:\theta}$ (\ie, $f^{\text{BLIP}}_{\theta}=E_{txt:\theta}$, and $\mathbf{c}=\mathrm{\hat{z}}_{cls}$), without incorporating additional modules. BLIP's text encoder inherently fuses image and text signals in its cross-attention layer, eliminating the need for a separate combining module. Notably, we freeze all vision encoders in our setup to ensure compatibility with existing image retrieval galleries and to enhance training efficiency.

\paragraph{Supervised / Pseudo Separated Contrastive Loss.} The training objective of CIR is to achieve strong alignment between the target image's embedding $\mathbf{x}$ (where $\mathbf{x}=\mathrm{\hat{x}}^t_{cls}$ for simplicity), and the composed embedding $\mathbf{c}$. On this purpose, we utilize HN-NCE \cite{HNNCE} loss for a given training batch $\mathcal{B\sim \mathit{D}}$ and $\mathcal{B' \sim \mathit{D'}}$, where $D'$ contains pseudo triplets. Additionally, to mitigate the impact of noise in pseudo triplets and ensure consistent contributions from supervised triplets, we compute a \textit{\textbf{t}arget-\textbf{c}omposed \textbf{c}ontrastive loss} (tcc) as: 

\begin{equation}
 \mathcal{L}_{tcc}(\mathcal{B}, \mathcal{B'}) =\mathcal{L}_{c}(\mathcal{B};\tau)+\mathcal{L}_{c} (\mathcal{B} \oplus \mathcal{B'};\tau) 
 \label{eqn:tcc}
\end{equation}

\noindent where $\mathcal{L}_c$ is defined as:

\begin{equation*}
\begin{split}
\mathcal{L}_{c}(\mathcal{B}) = - \sum_{i=1}^{n} \log \left[ \frac{e^{\mathbf{x}_i^T \mathbf{c}_i / \tau}}{\alpha \cdot e^{\mathbf{x}_i^T \mathbf{c}_i / \tau} + \sum_{j \neq i}^{\mathcal{B}} e^{\mathbf{x}_i^T \mathbf{c}_j / \tau} \cdot
 w_{\mathbf{x}_i,\mathbf{c}_j}} \right] \\ 
- \sum_{i=1}^{n} \log \left[ \frac{e^{\mathbf{x}_i^T \mathbf{c}_i / \tau}}{\alpha \cdot e^{\mathbf{x}_i^T \mathbf{c}_i / \tau} + \sum_{j \neq i}^{\mathcal{B}} e^{\mathbf{x}_i^T \mathbf{c}_j / \tau} \cdot w_{\mathbf{x}_j,\mathbf{c}_i}} \right].
\end{split}
\end{equation*}

\noindent where $\tau, \alpha$ denote hyper-parameters, and $\oplus$ denotes concatenation along the batch axis, and $w_{\mathbf{x}_i,\mathbf{c}_j}, w_{\mathbf{x}_j,\mathbf{c}_i}$ are set as in \cite{HNNCE}. This design facilitates the independent yet concurrent investigation of both supervised and semi-supervised CIR embedding spaces in a contrastive manner.

\label{sec:cir_training}
\begin{figure}[!t]
\centering
\includegraphics[width=0.9\linewidth]{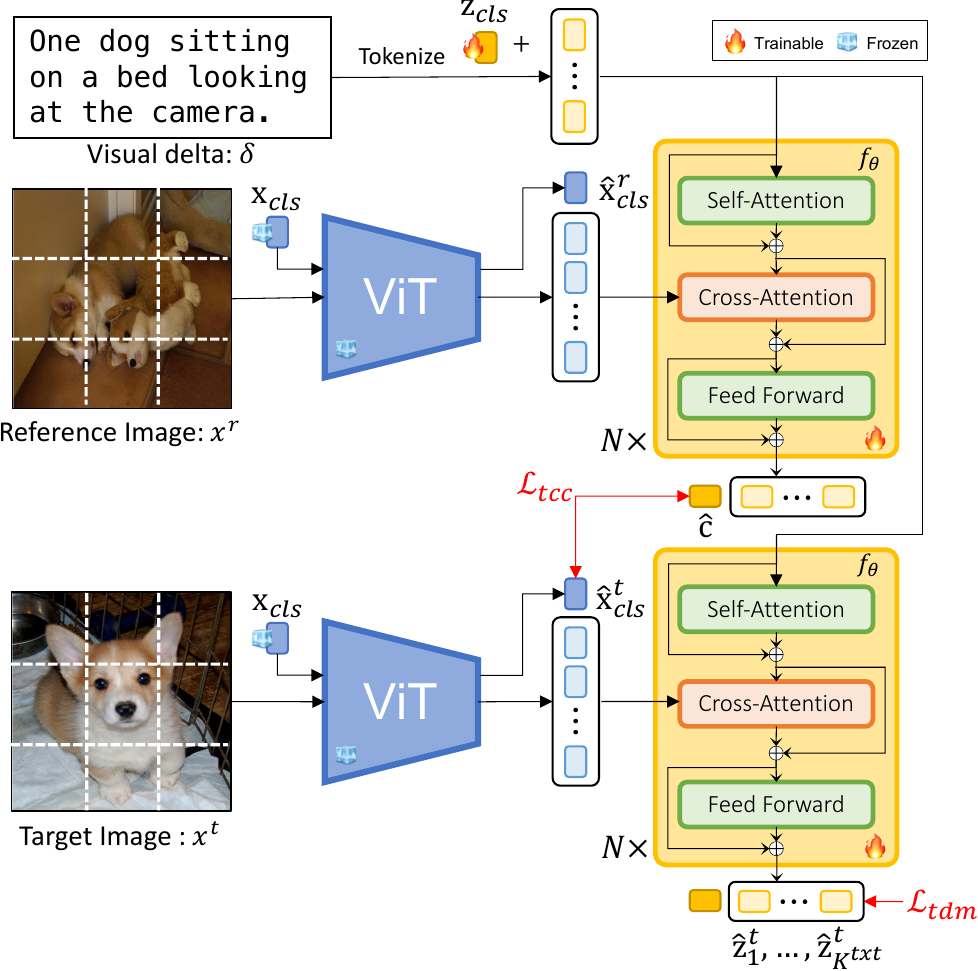}
\caption{Illustration of our proposed adaptation of the BLIP image-grounded text encoder for CIR. Both reference ($x^r$) and target image ($x^t$) patch tokens are processed by the text encoder ($f_{\theta}$).}
\label{fig:blip_training}
\end{figure}

\paragraph{Target-Delta Matching Loss.} In the case of BLIP's image-grounded text encoder structure, we introduce a new \textit{\textbf{t}arget-\textbf{d}elta \textbf{m}atching loss} (tdm). Our insight stems from the observation that, while all reference image patch tokens are considered in the cross-attention layer of the BLIP text encoder, target image patch tokens are overlooked when training solely with contrastive learning between cls token embeddings. As an example shown in Fig.~\ref{fig:blip_training}, the visual delta can be seen as a \textit{weakly correlated caption} to the target image. Thus, we aim to align the target image and visual delta to enable the BLIP text encoder to identify the image tokens related to textual input. The tdm loss is applied as:

\begin{equation}
\begin{split}
% \mathcal{L}_{\text{tdm}} = \frac{1}{K_{\text{txt}}} \sum^{K_{\text{txt}}}_{i=1} \mathcal{L}_{\text{cla}}(\mathrm{f_{\theta}}(\mathrm{\hat{x}}^i_{cls}, \mathrm{\hat{c}}^{j}), y^{ij}) \\
 % \mathcal{L}_{\text{tdm}} = \frac{1}{K_{\text{txt}}} \sum^{K_{\text{txt}}}_{i=1} \mathcal{L}_{\text{cla}}(\mathrm{\hat{z}}^t_i),
  \mathcal{L}_{\text{tdm}} = \mathbb{E}_{(x^t, \delta) \sim (D,D')}\mathrm{H}(\mathbf{y}^{tdm}, \mathbf{p}^{tdm}(x^t,\delta))
 % \mathcal{L}_{\text{cla}}(x, c) = \frac{1}{K_{\text{txt}}} \sum^{K_{\text{txt}}}_{i=1} \mathcal{L}_{\text{cla}}(\mathrm{\hat{z}}^t_i),
\label{eqn:tdm}
\end{split}
\end{equation}
\noindent where $\mathrm{H}$ is cross entropy, $\textbf{y}^{tdm}$ is a 2-dimensional one-hot vector label obtained through the hard negative mining process proposed in \cite{Align}, where a pair of image and text is configured as matched or not. $\mathbf{p}^{tdm} \in \mathbb{R}^2$ generates both positive and negative target-delta matching scores and is computed as:
\begin{equation*}
\mathbf{p}^{tdm}(x^t,\delta)=\sum\nolimits_{i=1}^{K^{txt}}p_{\theta}(\mathrm{\hat{z}}_i^t)/K^{txt},
\end{equation*}

\noindent where $\mathrm{\hat{z}}_{i}^t$ denotes the $i$-th token embedding in $f_\theta(x^t, \delta)$ and $p_{\theta}$ is a FC layer that outputs a 2-D vector. This loss ensures the text encoder fully processes the target image features, improving its understanding of the visual delta.

% We minimize the final loss function $\mathcal{L}$ with respect to the fusion function $f_{\theta}$:
% \begin{equation}
% \begin{split}
%  \mathcal{L} &=  \mathcal{L}_{tcc}(x^r,\delta,x^t) + \mathcal{L}_{tcc}(x'^r,\delta',x'^t) \\
%  & + \mathcal{L}_{tdm}(x^t,\delta) + \mathcal{L}_{tdm}(x'^t,\delta')   
% \end{split}
% \end{equation}

%% file: sec/4_experiment.tex
\section{Experiments}

Sec.~\ref{sec:setup} details the setup and VDG generation. Sec.~\ref{sec:quality_check} covers quality checks of visual deltas. Evaluations and baseline comparisons are in Sec.~\ref{sec:comparison}, and further analyses in Sec.~\ref{sec:further_analyses}.

\subsection{Setup}
\label{sec:setup}

\paragraph{Implementation Details.} %For VDG component, 
We utilize InstructBLIP's pretrained weights \cite{InstructBLIP}, which have been trained with both a ViT-G/14 \cite{ViT-G} and the Q-Former based on the Vicuna-13B model \cite{Vicuna}, without using prompts for Q-Former. For the stage 1 training described in Sec.~\ref{sec:VDG_training}, we employ 595K filtered image-text pairs from CC3M \cite{CC3M} provided by \cite{llava} to find alignment with our baseline LLM, LLaMA2-13B \cite{llama2}. In Stage 2, we implement instruction tuning with LoRA parameters \cite{LoRA}, following the fixed prompt as outlined in Fig.~\ref{fig:instruction_tuning}. Visual deltas are generated in an autoregressive manner, predicted based on the LLaMA2 vocabulary. For the CLIP-based CIR training, we select the ViT-L/14 model combined with a Combiner \cite{Combiner}. For the BLIP-based model \cite{BLIP}, we use a dedicated BLIP text encoder for image-text matching, paired with the ViT-L/16 model. Additional details can be found in the appendix.

% For vision projector part of VDG, we adopt the parameters of InstructBLIP \cite{InstructBLIP} with ViT-G/14 \cite{ViT-G} and Q-Former, which are aligned with Vicuna-13B \cite{Vicuna}. We choose LLaMA2-13B \cite{llama2} as our baseline LLM, and train the projection layer between vision projector and LLM for alignment with learning rate of 1e-3 for 1 epoch. We apply instructional tuning on additional parameters of LoRA \cite{LoRA} with the hyper-parameters as: $\alpha=16$, $\text{rank}=64$, $\text{dropout}=0.05$, where initial learning rate of 2e-4 with warmup for 100 iterations, reducing it to one-tenth when it reaches half of the total 10 epochs. For CLIP-based \cite{CLIP} CIR baseline, we choose ViT-L/14 model with Combiner \cite{Combiner}. For BLIP-based \cite{BLIP} CIR baseline, we utilize BLIP text encoder trained for image-text matching, with ViT-L/16 model. The entire training process is optimized with AdamW \cite{AdamW} with $\beta_1=0.9$, $\beta_2=0.99$ and a weight decay of 0.05. For image augmentation, basic random resized crop is applied with scale (0.8, 1.0) and ratio (0.9, 1.1). All models are trained with 8-NVIDIA A100 80GB GPUs as `bfloat16' type. The generation of visual delta from VDG is done autoregressively with the 0.2 temperature scaling on top 50 token predictions. 

\paragraph{Datasets for CIR Evaluation.} %In the realm of CIR, 
There are two standard benchmarks in CIR, one is CIRR \cite{CIRR} which deals with natural images, and the other is FashionIQ \cite{fashionIQ} which focuses on fashion domain images. Each presents unique challenges and datasets that help researchers push the boundaries of what's possible in CIR. Following the protocols utilized in benchmarks \cite{CIRR, fashionIQ}, we report the CIR results with recall scores at top K retrieval results (R@K), or results under collected subset ($\text{R}_{s}$@K). Specifically, \textit{CIRR} is configured with 4,351 subgroups (subsets), each containing six similar images, sourced from NLVR2 \cite{NLVR2}. For experimental purposes these groups are distributed into train (3,345 subgroups of 16,742 images), validation (503 subgroups of 2,265 images), and test (503 subgroups of 2,178 images) sets. \textit{FashionIQ} is divided into three categories of Dress, Shirt, and Toptee (Tops and Tees). The reference and target images are paired based on their category similarities. The 18,000 CIR triplets for training are pooled from 45,429 images of the training set, and 6,016 CIR triplets for the test are chosen from 15,415 images of the validation set.

\begin{figure}[!t]
\centering
\includegraphics[width=0.9\linewidth]{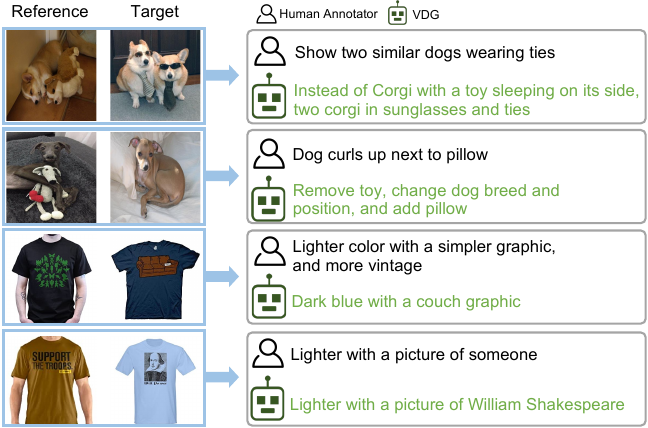}
\caption{Qualitative comparison on visual deltas, human \vs VDG on CIRR and FashionIQ datasets. For both natural and fashion domain images, VDG can produce informative visual deltas.}
\label{fig:qualitative_results}
\end{figure}

\paragraph{Visual Delta Generation.} To produce pseudo triplets, we expand our dataset sources to configure an \textit{auxiliary gallery} ($G'$) which is built upon the grouping strategy introduced in Sec. \ref{sec:pseudo_triplet_generation}, while \textit{excluding images that overlap} with the benchmark sets. Once the subgroups are constructed, we further filter them to avoid heavy overlap. In total, we draw upon 42,390 unique subgroups from NLVR2 \cite{NLVR2}, and 79,427 from COCO \cite{COCO}. Similarly, we build 27,957 individual subgroups from FashionIQ \cite{fashionIQ}, and 30,880 from DeepFashion \cite{DeepFashion}. We mark \textit{upper strophe} ($'$) to these datasets. For the semi-supervised settings, we randomly select 3,345 groups from $\text{NLVR2}'$ and $\text{COCO}'$ for the CIRR case, as well as 3,600 groups from $\text{FashionIQ}'$ and $\text{DeepFashion}'$. We denote these datasets used for semi-supervised CIR with the subscript ${se}$, (\eg, $\text{NLVR2}'_{se}$, $\text{COCO}'_{se}$). Fig.~\ref{fig:qualitative_results} shows the comparison between human-annotated visual deltas and those generated by the VDG in both natural and fashion domains. We observe that the VDG is effective in generating high-quality visual deltas -- additional results can be found in the appendix.

\subsection{Quality Check of VDG Responses} 
\label{sec:quality_check}

\input{tables/table1}

To assess the quality of VDG-generated visual deltas, we executed a series of experiments as outlined in Table \ref{table:Validation_VDG}. We use two backbones, Combiner, and BLIP with the tdm loss (denoted as $\text{BLIP}_{\text{tdm}}$), as our baselines for evaluation.

Initially, we compare the visual deltas generated by VDG with those annotated by humans. This is done by replacing the deltas in the training set with VDG-generated deltas for the same reference-target pairs. When comparing (a) and (b), we observe only a marginal drop in performance upon switching to VDG-generated deltas. This indicates that the deltas generated by VDG are as effective as those created by humans. More importantly, the improved performance observed when comparing (a) with (c) — which combines human-annotated and VDG-generated deltas — highlights the effectiveness of this hybrid approach as a robust data augmentation strategy to enhance CIR models.

\input{tables/table2}

\input{tables/table3}
To evaluate VDG's performance on new image pairs, we conduct experiments by substituting the human-annotated deltas in the validation set with deltas generated by VDG. The comparison of (a) and (d) reveals comparable performance levels, suggesting that deltas generated by VDG are as effective as human annotations. When we incorporate VDG-generated deltas into the training set (cases (e) and (f)), we achieve a notable performance improvement over the human-annotated validation set across all metrics, with the exception of a small reduction in R@1. For example, there is a 5.14\%p increase in R@5 between (b) and (e). The improvement observed may stem from the varied text descriptions by different human annotators, leading to inconsistencies in style and detail that potentially affect the distribution of visual differences between the annotated triplets. VDG offers a more consistent solution in generating visual deltas, thereby reducing errors associated with differences in human annotation. Our comprehensive testing demonstrates VDG’s reliability as an annotation tool for CIR.

\subsection{Comparison with Other Methods}
\label{sec:comparison}

Tables \ref{table:Comparisons_CIRR} and \ref{table:Comparisons_FashionIQ} present comparative evaluations between our method and other CIR approaches on CIRR and FashionIQ evaluation protocols. We categorize the experiments into two distinct groups. The first group, \textcolor{supervisedBase}{\textit{Seen}},  includes scenarios where the CIR model is trained with human-annotated training triplets. In contrast, the second group, \textcolor{unsupervisedBase}{\textit{Unseen}}, comprises scenarios in which the CIR model is trained without human-annotated triplets. Our implementations of Combiner and BLIP are denoted by \colorbox{grey}{grey box}. Specifically, methods under (a) are trained with supervised triplets, while those notated with the symbol ($\dagger$) are additionally augmented with VDG-generated visual deltas from supervised image pairs (\ie, the same setting as Human + VDG  in Table~\ref{table:Validation_VDG}). The methods in (b) have been pretrained on external datasets for CIR tasks, resulting in enhanced performances. In our semi-supervised setup (c), the models are trained with augmented supervised triplets and pseudo triplets sub-sampled from the auxiliary gallery as described in Secs.~\ref{sec:pseudo_triplet_generation} and~\ref{sec:setup}. Methods in (d) are trained with large-scale datasets with the intention of producing zero-shot performances. In (e), we only use pseudo triplets generated from the auxiliary gallery for training.

The results clearly demonstrate that our VDG implementation significantly improves the performance of both Combiner and $\text{BLIP}_{tdm}$ baselines. In (a), a detailed comparison shows that models augmented with VDG samples (with $\dagger$) consistently outperform their counterparts (without $\dagger$) in all recall metrics. Moreover, in (c), when we enhance our CIR baselines with additional pseudo triplets, there is a notable performance improvement. Specifically, when $\text{BLIP}_{tdm}$ is combined with the auxiliary gallery, it surpasses the previous state-of-the-art results in most recall metrics, showcasing the advantages of VDG. Additionally, it's important to highlight that our semi-supervised CIR method achieves these improvements with considerably fewer images than methods like CASE or CoVR, demonstrating its efficiency.

%We also compare the unseen case where a backbone does not see any human-annotated visual delta in (d) and (e).

In the unseen scenario, our implemented Combiner baseline not only outperforms Pic2Word, which uses the same CLIP backbone, but our $\text{BLIP}_{tdm}$ also significantly exceeds the former best-performing SEARLE or CoVR in most metrics. VDG's key advantage is its ability to generate visual deltas that align with the targeted domain, such as using COCO for natural images and DeepFashion for fashion images, by solely utilizing images without any accompanying captions. As a result, VDG significantly reduces the number of training samples required, yet achieves enhanced retrieval performances compared to zero-shot approaches like Pic2Word, CASE, or CoVR. This demonstrates that by focusing on similar domain, image-only datasets, we can substantially improve the efficacy of CIR model training.

% The reason for slightly lower performance than CASE for lower recall ranks is caused by choice of training objective, where we choose HN-NCE \cite{HNNCE} while CASE used \cite{Recallatk} which is specified for recall based retrieval system.
 % Due to its intrinsic design focused on identifying compositional relationships between image pairs, NLVR2 emerges as a superior candidate for generating additional visual deltas, closely mirroring the reference and target image dynamics inherent in CIR triplets.

 % *Combiner & 44.52 & 71.34 & 46.17 & 67.22 & 48.39 & 76.08 \\
 % *$\text{BLIP}_{tdm}$& 53.00 & 78.24 & 51.57 & 73.41 & 56.76 & 80.21 \\ \midrule
 % *Combiner & 35.99 & 63.36 & 38.96 & 60.84 & 42.02 & 67.92\\
 % *$\text{BLIP}_{tdm}$ & 49.13 & 74.76 & 50.05 & 71.30 & 53.24 & 78.74 \\ \midrule
 % *Combiner & & & & & & \\
 % *$\text{BLIP}_{tdm}$& 52.45 & 77.44 & 52.01 & 74.09 & 56.25 & 80.67 \\ \midrule
\input{tables/table4}

\begin{figure}[!t]
\centering
  \subcaptionbox{CIRR validation
  \label{fig:Recall_cirr}}{\includegraphics[height=2cm]{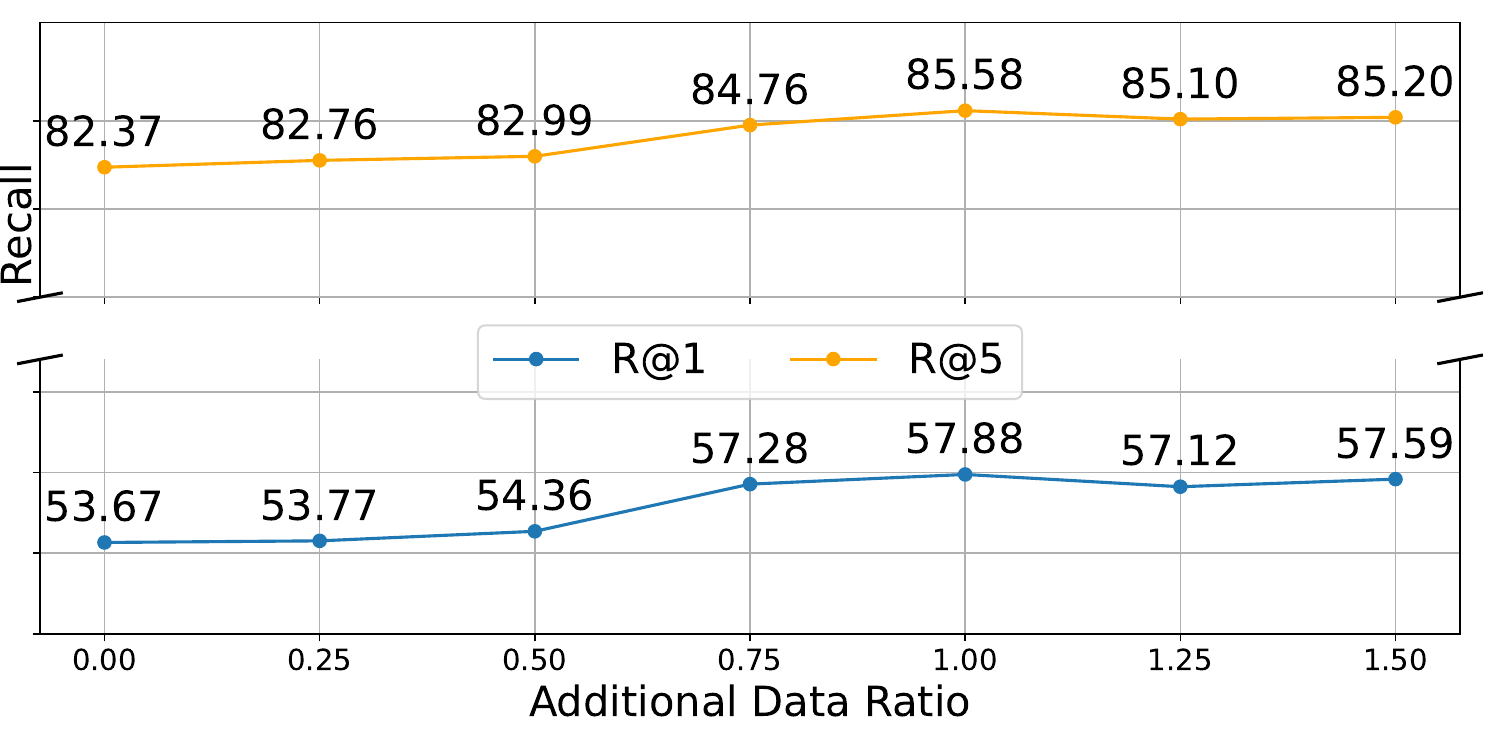}}
  \subcaptionbox{FashionIQ-Shirt
  \label{fig:Recall_fashion}}{\includegraphics[height=2cm]{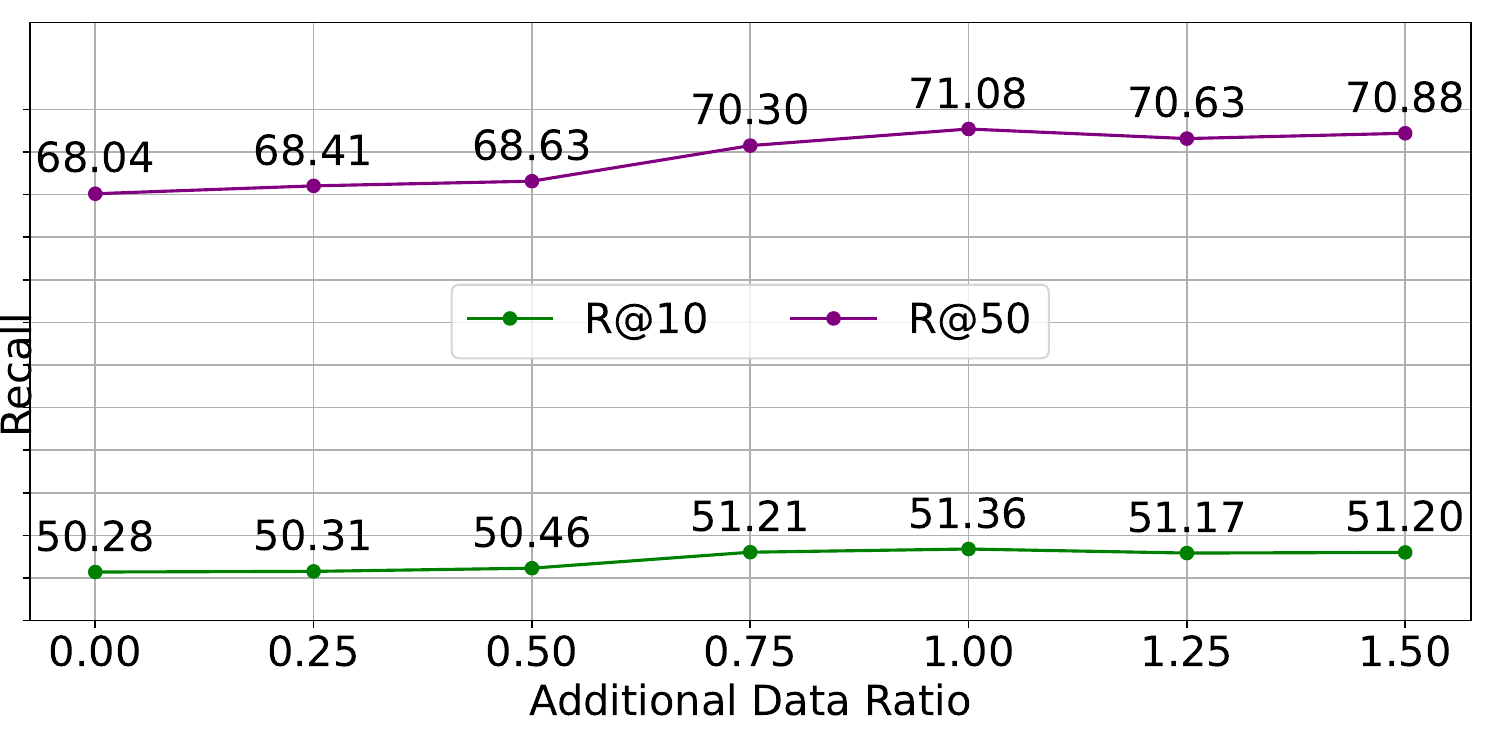}}
  \vspace{-2mm}
\caption{Analysis on the scale for \textcolor{supervisedBase}{\textit{(c) Supervised + Auxiliary Gallery with VDG}} of $\text{NLVR2}'_{se}$ and $\text{FashionIQ}'_{se}$. The x-axis denotes the ratio of additional images to training images.}
\label{fig:recall_scores}
\end{figure}

\input{tables/table5}

\subsection{Further Analyses}
\label{sec:further_analyses}

\noindent \textbf{Ablation study on Each Component.} To showcase the effectiveness of our proposed learning strategies, we conducted an ablation study and presented the findings in Table \ref{table:ablation}.  We consider three scenarios: (1) without \textit{grouping}, where we randomly sample images from the auxiliary gallery to create pairs; (2) without \textit{concat.}, where we remove the concatenation step used in Eqn. \ref{eqn:tcc} and use pseudo triplets only; and (3) without \textit{tdm loss}, where we exclude the tdm loss during training. The results indicate that each element sufficiently contributes to enhancing the performance.

\noindent \textbf{Impact of Generated Data Scale.} Figs. \ref{fig:Recall_cirr} and \ref{fig:Recall_fashion} demonstrate the effect of the size of VDG-generated data. In a semi-supervised setup, we find that the performance peaks when the number of pseudo-selected images approximates the size of the training set, and then saturates. This saturation might arise because additional pseudo triplets fail to represent the test sample distribution, particularly as we rigorously remove overlapping images with the test set when forming $G'$. For the unseen case as shown in Table \ref{table:ablation_scale}, performance is enhanced with an increase in VDG-generated data. This suggests that even in the absence of supervised triplets for CIR model training, a larger set of pseudo triplets not only expands the model’s comprehension of visual compositions but also aligns more closely with the distribution of supervised triplets. This alignment facilitates improved targeted domain retrieval performances, striking a balance between generalization and performance enhancement.

\begin{figure}[!t]
\centering
\includegraphics[width=0.92\linewidth]{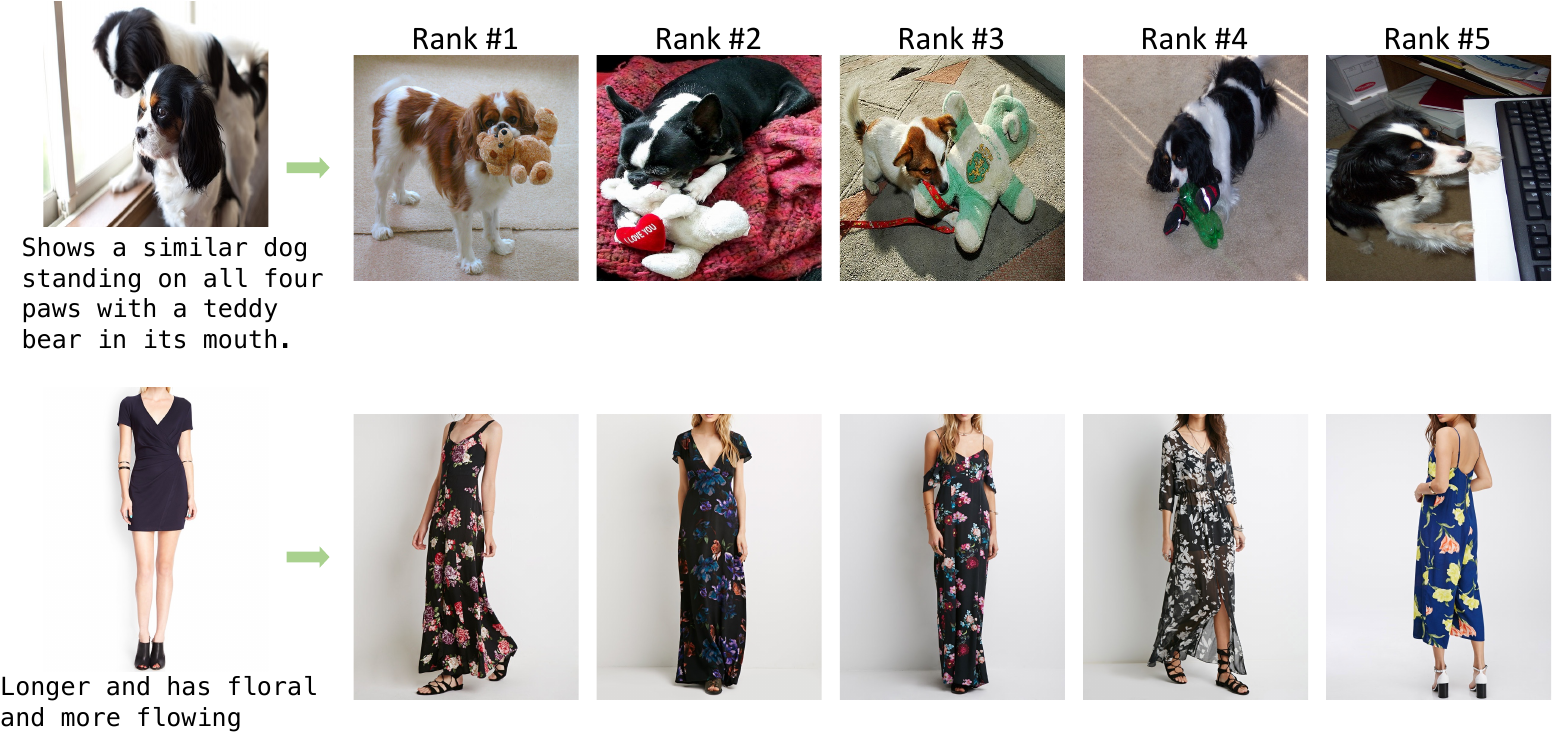}
\vspace{-2mm}
\caption{Retrieval results on auxiliary galleries, COCO, and DeepFashion. Actual user intents are used as text queries.}
\label{fig:retrieval_results}
\vspace{-1.5em}
\end{figure}

\noindent \textbf{Qualitative Retrieval Results.} We facilitate retrieval by providing user intents along with query images, drawing from auxiliary galleries. The retrieval results, as depicted in Fig. \ref{fig:retrieval_results}, demonstrate accurate and relevant image retrieval. Additional results are detailed in the appendix.

%% file: tables/table1.tex
\begin{table}[t]
\centering
\vspace{-2mm}
\caption{Retrieval results on \textit{CIRR validation set}. Human + VDG represents utilizing both human annotated and VDG-generated visual deltas when training the CIR model. $\text{BLIP}_{tdm}$ represents a model with target-delta matching loss. The best scores for each Val. set supervision are highlighted in \textbf{bold}.}
% By categorizing with Val. set supervision, 
\begin{adjustbox}{width=0.45\textwidth}
\begin{tabular}{c|c|l|c|c|c|c}
\toprule
% \multirow{2}{*}{\makecell{Training Set \\ Supervision}}  &\multirow{2}{*}{\makecell{Val. Set \\ Supervision}} & \multicolumn{1}{c|}{\multirow{2}{*}{Baseline}} & \multicolumn{4}{c}{R@K} \\ \cmidrule{4-7}  &\multicolumn{1}{c|}{} &\multicolumn{1}{c|}{} & K=1   & K=5  & K=10  & K=50   \\ \midrule 
{\makecell{Training Set \\ Supervision}}  &{\makecell{Val. Set \\ Supervision}} & Baseline & R@1 & R@5  & R@10  & R@50   \\ \midrule 
\multicolumn{1}{l|}{\multirow{2}{*}{(a) Human}} & \multirow{2}{*}{Human}
& Combiner & 37.98 & 71.49 & 82.52 & 95.29 \\
% & & BLIP & 52.69 & 82.01 & 89.45 & 97.44 \\ 
& & $\text{BLIP}_{tdm}$ & 53.17 & 82.09 & 89.81 & 97.54\\\midrule
\multicolumn{1}{l|}{\multirow{2}{*}{(b) VDG}} & \multirow{2}{*}{Human}
& Combiner & 35.47 & 68.29 & 80.29 & 94.31 \\
% & & BLIP & 49.94 & 79.26 & 87.78 & 96.65 \\
& & $\text{BLIP}_{tdm}$  & 50.16 & 80.03 & 87.78 & 96.75\\ \midrule
\multicolumn{1}{l|}{\multirow{2}{*}{(c) Human + VDG}} & \multirow{2}{*}{Human}
& Combiner & 39.11 & 73.02 & 84.41 & 95.96 \\
% & & BLIP & 53.41 & 81.61 & 89.55 & 97.56 \\
& & $\text{BLIP}_{tdm}$  & \textbf{53.67} & \textbf{82.99} & \textbf{89.97} & \textbf{97.84} \\ \midrule
\multicolumn{1}{l|}{\multirow{2}{*}{(d) Human}} & \multirow{2}{*}{VDG}
& Combiner & 38.96 & 73.31 & 84.36 & 96.22 \\
% & & BLIP & 51.59 & 83.31 & 91.41 & 98.01 \\ 
& & $\text{BLIP}_{tdm}$  & 51.64 & 83.33 & 91.39 & 97.90\\ \midrule
\multicolumn{1}{l|}{\multirow{2}{*}{(e) VDG}} & \multirow{2}{*}{VDG}
& Combiner & 41.59 & 77.57 & 87.35 & 97.11 \\
% & & BLIP & 52.67 & 85.17 & 92.20 & 98.54 \\ 
& & $\text{BLIP}_{tdm}$  & 52.69 & 85.17 & 92.37 & 98.59\\ \midrule
\multicolumn{1}{l|}{\multirow{2}{*}{(f) Human + VDG}} & \multirow{2}{*}{VDG}
& Combiner & 41.47 & 77.90 & 87.42 & 97.30 \\
% & & BLIP & 52.81 & 84.88 & 91.94 & 98.47 \\ 
& & $\text{BLIP}_{tdm}$  & \textbf{52.95} & \textbf{85.27} & \textbf{92.54} & \textbf{98.61} \\ 
\bottomrule
\end{tabular}
\end{adjustbox}
\label{table:Validation_VDG}
\end{table}

%% file: tables/table2.tex
\begin{table}[!t]
\centering
\vspace{-2mm}
\caption{Retrieval results on \textit{CIRR test set}. * denotes our baselines, $\dagger$ denotes VDG generated visual deltas are applied to augment original training set. We categorize into two distinct groups: one is \textcolor{supervisedBase}{\textit{Seen: Supervised / Supervised + External / Supervised + Aux. with VDG}} which utilize human-annotated visual delta for CIR model training, and the other is \textcolor{unsupervisedBase}{\textit{Unseen: Zero-shot / Aux. with VDG}} which does not utilize human-annotated visual delta for CIR model training. Within each category, best viewed with \textbf{bold}.}
\begin{adjustbox}{width=0.49\textwidth}
\begin{tabular}{l|c|c|c|c|c|c|c}
\toprule
 % \multicolumn{1}{c|}{\multirow{2}{*}{Method}} & \multicolumn{4}{c|}{R@K} & \multicolumn{3}{c}{$\text{R}_{subset}$@K} \\ \cmidrule{2-8} 
 %                                                 & K=1   & K=5  & K=10  & K=50   & K=1  & K=2  & K=3  \\ \midrule
Method & R@1 & R@5 & R@10 & R@50 & {$\text{R}_{s}$@1} & {$\text{R}_{s}$@2} & {$\text{R}_{s}$@3} \\ \midrule
                           \multicolumn{8}{c}{\textcolor{supervisedBase}{\textit{(a) 
 Supervised}}} \\ \midrule

 % TIRG \cite{TIRG}   & 14.61 & 48.37 & 64.08 & 90.03 & 22.67  & 44.97  &  65.14 \\
 % MAAF \cite{MAAF}  & 10.31 & 33.03 & 48.30 & 80.06 & 21.05 &  41.81 &  61.60  \\
 % MAAF-RP \cite{MAAF} & 10.22 & 33.32 & 48.68 & 81.84 & 21.41 & 42.17 &  61.60 \\
 ARTEMIS \cite{ARTEMIS} & 16.96 &  46.10 & 61.31 & 87.73 & 39.99 & 62.20 & 75.67 \\
 CIRPLANT \cite{CIRPLANT} & 19.55 & 52.55 & 68.39 & 92.38 & 39.20 & 63.03 & 79.49 \\
 Combiner \cite{Combiner} & 33.59 & 65.35 & 77.35 & 95.21 & 62.39  & 81.81  & 92.02 \\
 CASE \cite{CASE} & 48.00 &  79.11 & 87.25 & 97.57 & 75.88 & 90.58 & 96.00\\
 CoVR \cite{CoVR} & 48.84  & 78.05 & 86.10 & 94.19 & 75.78 & 88.22 & 92.80 \\ 
\rowcolor{oursColor} Combiner & 34.39 & 66.22 & 76.58 & 91.04 & 68.55 & 86.36 & 93.98 \\
\rowcolor{oursColor} $\text{Combiner}^{\dagger}$ & 36.91 & 69.21 & 79.54 & 92.04 & 70.00 & 87.45 & 94.39 \\
\rowcolor{oursColor} $\text{BLIP}_{tdm}$ & 48.94 & 77.83 & 86.15 & 94.17 & 75.71 & 89.71 & 95.81 \\ 
\rowcolor{oursColor} $\text{BLIP}_{tdm}^{\dagger}$ & 49.08 & 78.98 & 86.89 & 94.24 & 76.18 & 90.62 & 95.86 \\ \midrule
 \multicolumn{8}{c}{\textcolor{supervisedBase}{\textit{(b) Supervised + External Dataset for Pretraining}}} \\ \midrule
 CASE + LasCo.Ca. \cite{CASE} & 49.35 & 80.02 &  \textbf{88.75} & \textbf{97.47} & 76.48 & 90.37 & 95.71 \\
 CoVR + WebVid \cite{CoVR} & 49.69 & 78.60 & 86.77 & 94.31 & 75.01 &  88.12 &  93.16 \\ \midrule
\multicolumn{8}{c}{\textcolor{supervisedBase}{\textit{(c) Supervised + Auxiliary Gallery with VDG}}} \\ \midrule
\rowcolor{oursColor} $\text{Combiner}^{\dagger}$ + $\text{COCO}'_{se}$ & 38.77 & 69.25 & 79.21 & 91.52 & 71.25 & 87.49 & 94.34 \\
\rowcolor{oursColor}  $\text{Combiner}^{\dagger}$ + $\text{NLVR2}'_{se}$ & 38.89 & 69.84 & 79.41 & 91.18 & 71.92 & 87.89 & 94.29 \\
\rowcolor{oursColor} $\text{BLIP}_{tdm}^{\dagger}$ + $\text{COCO}'_{se}$ & 49.37 & 78.12 & 85.52 & 93.74 & 76.68 & 90.46 & 96.05  \\ 
\rowcolor{oursColor}  $\text{BLIP}_{tdm}^{\dagger}$ + $\text{NLVR2}'_{se}$ & \textbf{50.96} & \textbf{80.15} & 86.86 & 94.46 & \textbf{77.45} & \textbf{90.65} & \textbf{96.10} \\ \bottomrule \toprule
 \multicolumn{8}{c}{\textcolor{unsupervisedBase}{\textit{(d) Zero-shot}}} \\ \midrule
 Pic2Word by CC3M \cite{Pic2word} & 23.90  & 51.70   & 65.30   & 87.80 & -	& - & -  \\
  SEARLE by ImageNet \cite{SEARLE}  & 24.22 &  52.41 & 66.29 & 88.63  & 53.71 & 74.63 &  87.61 \\ 
 CASE by LasCo.Ca. \cite{CASE} & 35.40 & 65.78 & 78.53 & \textbf{94.63}  & 64.29 &  82.66 &  91.61 \\
 CoVR by WebVid \cite{CoVR}  & 38.48 &  66.70 & 77.25 & 91.47  &69.28  & 83.76 &  91.11 \\ \midrule
 \multicolumn{8}{c}{\textcolor{unsupervisedBase}{\textit{(e) Auxiliary Gallery with VDG}}} \\ \midrule
\rowcolor{oursColor} Combiner by $\text{COCO}'$  & 26.80 & 54.05 & 65.30 & 83.88 & 67.28 & 85.42 & 92.99 \\
\rowcolor{oursColor}  Combiner by $\text{NLVR2}'$  & 31.57 & 61.37 & 72.10 & 88.94 & 67.98 & 86.18 & 93.18 \\
\rowcolor{oursColor} $\text{BLIP}_{tdm}$ by $\text{COCO}'$  & 43.49 & 72.07 & 81.59 & 93.21 & 72.36 & \textbf{87.98} & 94.72  \\ 
\rowcolor{oursColor} $\text{BLIP}_{tdm}$ by $\text{NLVR2}'$  & \textbf{45.74} & \textbf{75.01} & \textbf{82.52} & 93.13 & \textbf{72.60} & 87.90 & \textbf{94.77}  \\ 

\bottomrule
\end{tabular}
\end{adjustbox}
\label{table:Comparisons_CIRR}
\end{table}

%% file: tables/table3.tex
\begin{table}[!t]
\centering
\vspace{-2mm}
\caption{Retrieval results on \textit{FashionIQ validation set}. We utilize the same notations and same categorizations as Table \ref{table:Comparisons_CIRR}.}
\begin{adjustbox}{width=0.49\textwidth}
\begin{tabular}{l|c|c|c|c|c|c}
\toprule
  \multicolumn{1}{c|}{\multirow{2}{*}{Method}} & \multicolumn{2}{c|}{Dress} & \multicolumn{2}{c|}{Shirt} & \multicolumn{2}{c}{Toptee}  \\ \cmidrule{2-7} 
                                  & R@10   & R@50  & R@10  & R@50   & R@10  & R@50 \\ \midrule
\multicolumn{7}{c}{\textcolor{supervisedBase}{\textit{(a) Supervised}}} \\ \midrule
 % JVSM \cite{JVSM} & 10.70 & 25.90 & 12.00  & 27.10 & 13.00  & 26.90 \\
 % CIRPLANT \cite{CIRPLANT} & 17.45 &  40.41  & 17.53 &  38.81 &  61.64 & 45.38 	  \\
 % Val w/GloVe \cite{GloVe} & 22.53 &  44.00 & 22.38 & 44.15 & 27.53 & 51.68 	  \\
 % MAAF \cite{MAAF} & 23.80 & 48.60 & 21.30 &  44.20 &  27.90 & 53.60 	  \\
 % CosMo \cite{CosMo}  & 25.64 & 50.30 & 24.90 & 49.18 & 29.21 & 57.46 	  \\
 ARTEMIS \cite{ARTEMIS} & 27.16 & 52.40 & 21.78 &  43.64 & 29.20 & 53.83  \\
 DCNet \cite{DCNet} & 28.95 & 56.07 & 23.95 & 47.30 & 30.44 & 58.29 \\
 FashionVLP \cite{FashionVLP} & 32.42 &  60.29 & 31.89 & 58.44 & 38.51 & 68.79  \\
 Combiner \cite{Combiner}   & 31.63 & 56.67 & 36.36 &  58.00 & 38.19 & 62.42  \\ 
 CoVR \cite{CoVR} & 43.51 & 67.94 & 48.28 & 66.68 & 51.53 & 73.60  \\
\rowcolor{oursColor} *Combiner & 31.95 & 55.05 & 39.21 & 56.82 & 38.55 & 62.16 \\
\rowcolor{oursColor}  *$\text{Combiner}^{\dagger}$ & 35.40 & 59.99 & 42.30 & 61.63 & 43.09 & 66.96 \\
\rowcolor{oursColor} *$\text{BLIP}_{tdm}$ & 44.87 & 66.83 & 49.61 & 66.93 & 50.54 & 72.26 \\
\rowcolor{oursColor} *$\text{BLIP}_{tdm}^{\dagger}$ & 46.90 & 68.86 & 50.28 & 68.04 & 52.73 & 74.45 \\\midrule
 \multicolumn{7}{c}{\textcolor{supervisedBase}{\textit{(b) Supervised + External Dataset for Pretraining}}} \\ \midrule
 CASE + LasCo.Ca. \cite{CASE} & 47.44 & 69.36  & 48.48 & 70.23 & 50.18 & 72.24   \\
 CoVR + WebVid \cite{CoVR} & 44.55 & 69.03 & 48.43 & 67.42 & 52.60 & 74.31  \\ \midrule
\multicolumn{7}{c}{\textcolor{supervisedBase}{\textit{(c) Supervised + Auxiliary Gallery with VDG}}}  \\ \midrule
\rowcolor{oursColor}  $\text{Combiner}^{\dagger}$ + $\text{DeepFashion}'_{se}$ & 35.50 & 60.09 & 42.54 & 62.86 & 43.27 & 67.86 \\
\rowcolor{oursColor} $\text{Combiner}^{\dagger}$ + $\text{FashionIQ}'_{se}$ & 36.30 & 60.19 & 43.98 & 62.27 & 44.33 & 68.06 \\
\rowcolor{oursColor}  $\text{BLIP}_{tdm}^{\dagger}$ + $\text{DeepFashion}'_{se}$ & 47.10 & 69.10 & 49.95 & 69.96 & \textbf{53.90} & 74.35 \\
\rowcolor{oursColor} $\text{BLIP}_{tdm}^{\dagger}$ + $\text{FashionIQ}'_{se}$& \textbf{47.89} & \textbf{69.81} & \textbf{51.36} & \textbf{71.08} & 53.29 & \textbf{74.65} \\ \bottomrule \toprule
 \multicolumn{7}{c}{\textcolor{unsupervisedBase}{\textit{(d) Zero-shot}}} \\ \midrule
 Pic2Word by CC3M \cite{Pic2word} \  & 20.00 & 40.20 &  26.20 & 43.60    & 27.90  & 47.40 \\
  SEARLE by ImageNet \cite{SEARLE}  & 20.32  & 43.18 & 27.43  & 45.68  & 29.32 & 50.17 \\
 CoVR by WebVid \cite{CoVR}  & 21.95  & 39.05 & 30.37  & 46.12  & 30.78 & 48.73 \\\midrule
 \multicolumn{7}{c}{\textcolor{unsupervisedBase}{\textit{(e) Auxiliary Gallery with VDG}}} \\ \midrule
\rowcolor{oursColor}  Combiner by $\text{DeepFashion}'$& 23.30 & 46.36 & 30.86 & 49.02 & 31.87 & 51.96 \\
 \rowcolor{oursColor}   Combiner by $\text{FashionIQ}'$ & 28.26 & 51.46 & 32.58 & 51.28 & 34.88  & 55.79 \\
\rowcolor{oursColor} $\text{BLIP}_{tdm}^{\dagger}$ by $\text{DeepFashion}'$ & 32.67  & 54.39 & 35.48  & 55.05 & 39.47 & 59.92 \\ 
 \rowcolor{oursColor} $\text{BLIP}_{tdm}^{\dagger}$ by $\text{FashionIQ}'$ & \textbf{37.48}  & \textbf{58.70} & \textbf{37.29}  & \textbf{57.11}  & \textbf{42.12} & \textbf{62.32} \\ 

\bottomrule
\end{tabular}
\end{adjustbox}
\label{table:Comparisons_FashionIQ}
\end{table}

%% file: tables/table4.tex
\begin{table}[!t]
\centering
\vspace{-2mm}
\caption{Ablation results on $\text{BLIP}_{tdm}$ baseline for \textit{CIRR validation set} with $\text{NLVR2}'_{se}$ as auxiliary gallery. Best viewed with \textbf{bold}.}
\vspace{-2mm}
\begin{adjustbox}{width=0.38\textwidth}
\begin{tabular}{ccc|c|c|c|c}
\toprule
\textit{grouping}  &\textit{concat.} & \textit{tdm loss} & R@1 & R@5  & R@10  & R@50 \\ \midrule 
& & & 52.45 & 81.18 & 88.62,  & 96.88 \\
\checkmark & & & 55.66 & 83.45 & 91.21 & 97.33 \\ 
\checkmark & \checkmark & & 56.28 & 84.39 & 91.41 & 97.92\\
\checkmark & \checkmark & \checkmark & \textbf{57.88} & \textbf{85.58} & \textbf{93.21} & \textbf{98.33}\\

\bottomrule
\end{tabular}
\end{adjustbox}
\label{table:ablation}
\vspace{-2mm}
\end{table}

%% file: tables/table5.tex
\begin{table}[!t]
\centering
\vspace{-4mm}
\caption{Analysis on the scale for \textcolor{unsupervisedBase}{\textit{(e) Auxiliary Gallery with VDG}}. The dataset is scaled from one eighth (1/8) to the full set (1). Evaluated on CIRR Test set.}
\vspace{-2mm}
\begin{adjustbox}{width=0.48\textwidth}
\begin{tabular}{c|cccc||c|cccc}
\toprule
R@1 & 1/8 & 1/4  & 1/2  & 1 & R@10 & 1/8 & 1/4  & 1/2  & 1 \\ \midrule 
COCO & 	42.15 & 42.89 & 43.10  & \textbf{43.49} & COCO & 79.81 & 81.01 & 81.16  & \textbf{81.59} \\ 
NLVR2 & 44.31 & 44.99 & 45.13 & \textbf{45.74}  & NLVR2 & 82.48 & 82.75 & 82.43 & \textbf{82.52}  \\ 

\bottomrule
\end{tabular}
\end{adjustbox}
\label{table:ablation_scale}
\vspace{-1.5em}
\end{table}

%% file: sec/5_conclusion.tex
\blfootnote{Acknowledgment: This work was partially supported by Institute of Information \& communications Technology Planning \& Evaluation (IITP) grant funded by the Korea government (MSIT) (No. 2019-0-00079,  Artificial Intelligence Graduate School Program(Korea University)).} 

\section{Conclusion}

In this study, we investigate a novel semi-supervised learning approach in the context of Composed Image Retrieval (CIR). Our findings reveal that integrating human-annotated data with pseudo triplets generated by the Visual Delta Generator (VDG) significantly enhances the generalization capacity of CIR models. The VDG approach not only streamlines the generation of visual deltas but also emerges as a cost-efficient and effective alternative to extensive human annotation. This work paves the way for future research in semi-supervised learning and showcases VDG as a promising direction for advancing CIR systems.

%% file: sec/X_suppl.tex
\clearpage
\setcounter{page}{1}
\maketitlesupplementary

\section{Reproduction Guide}
\label{sec:reproduction_guide}

\subsection{More Implementation details}

\paragraph{VDG: Alignment.} For stage 1 in Sec. \ref{sec:VDG_training}, the focus is on training the projection layer in the vision projector and our baseline LLM, LLaMA2-13B, to achieve alignment. Training is executed over a single epoch with a learning rate of \(1 \times 10^{-3}\), batch size 64 per GPU.

\paragraph{VDG: Instructional Tuning.} For stage 2 in Sec. \ref{sec:VDG_training}, additional LoRA parameters ($\theta_{lora}$) are applied, configured with \(\alpha=16\), \(\text{rank}=64\), and \(\text{dropout}=0.05\). The tuning process begins at a learning rate of \(2 \times 10^{-4}\), incorporating a warm-up phase over 100 iterations. The learning rate is reduced to one-tenth after reaching half of the total 10 epochs, batch size 8 per GPU.

\paragraph{VDG: Augmentation.} For VDG training, we simply apply basic random resized cropping to our images. This involves adjusting the scale of the images between 0.8 and 1.0 and their aspect ratios between 0.9 and 1.1.

\paragraph{VDG: Visual Delta Generation}
The generation of visual deltas is conducted autoregressively, applying a temperature scaling of 0.2 to the top 50 token predictions.

\paragraph{CIR: Hyper-parameters} For Eqn. \ref{eqn:tcc}, we set the hyper-parameters as $\tau=0.01,\alpha=1.0,\beta=0.0$ for CLIP-based Combiner, which makes loss as same as standard contrastive loss, and $\tau=0.01,\alpha=1.0,\beta=0.5$ for BLIP baseline.

\paragraph{CIR: Augmentation.} CIR model training employs a standard data augmentation pipeline to enhance robustness. We start with a random resized crop, adjusting the scale of the images between 0.5 and 1.0. Further, a random horizontal flip, and random adjustments to image contrast, brightness, and sharpness are applied. We also incorporate different perspectives and angles of images by modifying translation and rotation. 

\paragraph{CIR: Model Training.} Batch size is set as 64 per GPU for CIR model training. The CIR models begin with an initial learning rate of $1e-4$, which follows a cosine decay schedule to zero for 6 and 10 epochs for BLIP and CLIP baselines, separately.

\paragraph{Model Training and Optimization}
All models are optimized using AdamW optimizer \cite{AdamW}, with \(\beta_1=0.9\), \(\beta_2=0.99\), and a consistent weight decay of 0.05. Training is performed on 8 NVIDIA A100 80GB GPUs using \textit{bfloat16} precision.

\subsection{Training procedure}

We provide a training procedure for VDG in Algorithms \ref{algorithm_alignment} and \ref{algorithm_instruction}, as well as a semi-supervised learning approach for the CIR model in Algorithm \ref{algorithm_semi_supervised_CIR}. We set the same batch size for $\mathcal{B}$ and $\mathcal{B}'$.

\begin{algorithm}[!h]
\caption{Stage 1 - Alignment of VDG}
\begin{algorithmic}[1]
\STATE Initialize $\theta_{proj}$
\STATE \textbf{Input}: $D$ - Filtered image-text pairs from CC3M
\STATE \textbf{Input}: $P$ - Set of prompts from LLaVA \cite{llava}
\STATE $\ell_{align} \leftarrow {\mathcal{L}_{LLM}}$ with $\mathcal{B} \sim D, p \sim P$
\STATE $\theta_{proj} \leftarrow{\theta_{proj} - \gamma \frac{\partial \ell_{align}}{\partial \theta_{proj}}}$
\ENSURE Updated $\theta_{proj}$
\end{algorithmic}
\label{algorithm_alignment}
\end{algorithm}

\begin{algorithm}[!h]
\caption{Stage 2 - Instruction Tuning of VDG}
\begin{algorithmic}[1]
\STATE Load $\theta_{proj}$ from Stage 1, initialize $\theta_{lora}$
\STATE \textbf{Input}: $D$ - Train set of CIRR or FashionIQ
\STATE \textbf{Input}: $p_{inst}$ - prompt from Fig. \ref{fig:instruction_tuning}
\STATE $\ell_{inst} \leftarrow {\mathcal{L}_{LLM}}$ with $\mathcal{B} \sim D, p_{inst}$
\STATE $\theta_{proj} \leftarrow{\theta_{proj} - \gamma \frac{\partial \ell_{inst}}{\partial \theta_{proj}}}$
\STATE $\theta_{lora} \leftarrow{\theta_{lora} - \gamma \frac{\partial \ell_{inst}}{\partial \theta_{lora}}}$
\ENSURE Updated $\theta_{proj}, \theta_{lora}$
\end{algorithmic}
\label{algorithm_instruction}
\end{algorithm}

\begin{algorithm}[!h]
\caption{Semi-supervised CIR training}
\begin{algorithmic}[1]
\STATE Load $E_{img}, f_{\theta}$
\STATE \textbf{Input}: $D$ - Train set of CIRR or FashionIQ (additional visual deltas are applied with VDG)
\STATE \textbf{Input}: $D'$ - Pseudo triplet generated from auxiliary gallery with VDG
\STATE $\ell_{tcc} \leftarrow {\mathcal{L}_{tcc}}$ with $\mathcal{B} \sim D, \mathcal{B}' \sim D'$
\STATE $\ell_{tdm} \leftarrow {\mathcal{L}_{tdm}}$ with $\mathcal{B} \sim D, \mathcal{B}' \sim D'$ (BLIP only)
\STATE $f_{\theta} \leftarrow{f_{\theta} - \gamma \frac{\partial (\ell_{tcc}+\ell_{tdm})}{\partial f_{\theta}}}$
\ENSURE Updated $f_{\theta}$
\end{algorithmic}
\label{algorithm_semi_supervised_CIR}
\end{algorithm}

\section{Further Analysis}

\begin{table}[!t]
\centering
\caption{Detailed dataset configurations for auxiliary galleries and CC3M-Filtered for alignment training (Stage 1 in Sec. \ref{sec:Method}). `\#' denotes the number of images in the dataset. `\#' ref-tar pairs denotes the number of unique reference-target image pairs.}
\begin{adjustbox}{width=0.33\textwidth}
\begin{tabular}{l|c|c}
\toprule
Dataset & \# images & \# ref-tar pairs \\ \midrule 
$\text{NLVR2}'$ & 63,788 & 152,604 \\
$\text{COCO}'$ & 102,436 & 285,939  \\
$\text{FashionIQ}'$ & 33,994 & 100,647  \\
$\text{DeepFashion}'$ & 38,237 & 111,168  \\
CC3M-Filtered & 595,375 & -  \\
%1,431,135
\bottomrule
\end{tabular}
\end{adjustbox}
\label{table:detailed_configuration_dataset}
\end{table}

\begin{table}[!t]
\centering
\caption{Ablation study on VDG for CIRR validation set.}
\begin{adjustbox}{width=0.45\textwidth}
\begin{tabular}{l|c|c|c|c}
\toprule
Methods & R@1 & R@5  & R@10  & R@50   \\ \midrule 
 Our Final Model & 50.16 & 80.03 & 87.78 & 96.75\\
 (1) LLaMA2-7B & 50.04 & 79.29 & 86.94 & 95.86 \\
 (2) LLaMA2-13B-chat & 50.23 & 79.67 & 86.96 & 96.03 \\
 (3) LoRA rank=32 & 49.03 & 79.12 & 86.80 & 96.24 \\
 (4) LoRA rank=128 & 48.94 & 79.05 & 86.96 & 95.86 \\
 (5) Q-Former: BLIP-2 & 49.63 & 78.76 & 86.68 & 95.72 \\
\bottomrule
\end{tabular}
\end{adjustbox}
\label{table:design_choice}
\end{table}

\begin{table}[!t]
\centering
\caption{Experiment results on the CIRR test set using different scales of the CC3M-Filtered Dataset ($\sim$1.4M pseudo CIR triplets). The dataset scales from one eight (1/8) to the full set (1).}
\begin{adjustbox}{width=0.32\textwidth}
\begin{tabular}{l|c|c|c|c}
\toprule
Ratio & R@1 & R@5  & R@10  & R@50   \\ \midrule 
 1/8 & 44.43 & 73.18 & 82.36 & 92.28\\
 1/4 & 45.34 & 73.90 & 82.82 & 93.21 \\
 1/2 & 45.37 & 74.24 & 83.06 & 93.25 \\
 1 & 45.42 & 74.55 & 83.28 & 93.32 \\
\bottomrule
\end{tabular}
\end{adjustbox}
\label{table:large_scale_exp}
\end{table}

\paragraph{Data statistics.} We provide detailed configuration of auxiliary gallery used for experiments in Table \ref{table:detailed_configuration_dataset}.

\paragraph{Design Choice.} We investigate several configuration options, including: (a) the model size of the LLM, (b) the type of LLM used, (c) the rank of LoRA, and (4) the type of Q-Former. Based on our final model in the first row, we change the designated component in each row. We explore these options in Table \ref{table:design_choice} to assess their impact on CIR performance. Specifically, for (1) we experiment with the LLaMA2-7B model. For (2), we opt for the chat-bot style tuned LLaMA2-13B-chat model. Regarding (3) and (4), we experiment with varying the rank at 32 and 128, noting that our baseline is 64. Finally, for (5), we employ the Q-Former from BLIP-2, which is in line with FlanT5-XXL \cite{FlanT5}, rather than using the InstructBLIP one. It is important to note that our evaluations indicate that these options do not significantly affect performance. This underscores the general applicability and robustness of our VDG, demonstrating its effectiveness across a variety of configurations.

\paragraph{Larger Scale Experiment.} We further configure 1,431,135 pseudo triplets with the CC3M-Filtered dataset to explore the scalability of VDG to larger datasets and show the results in Table \ref{table:large_scale_exp}. It appears that as the dataset size increases, there's a gradual improvement in the recall metrics, suggesting that using more data improves the model's ability to retrieve relevant results. 

\begin{figure*}[!t]
\centering
\includegraphics[width=0.99\linewidth]{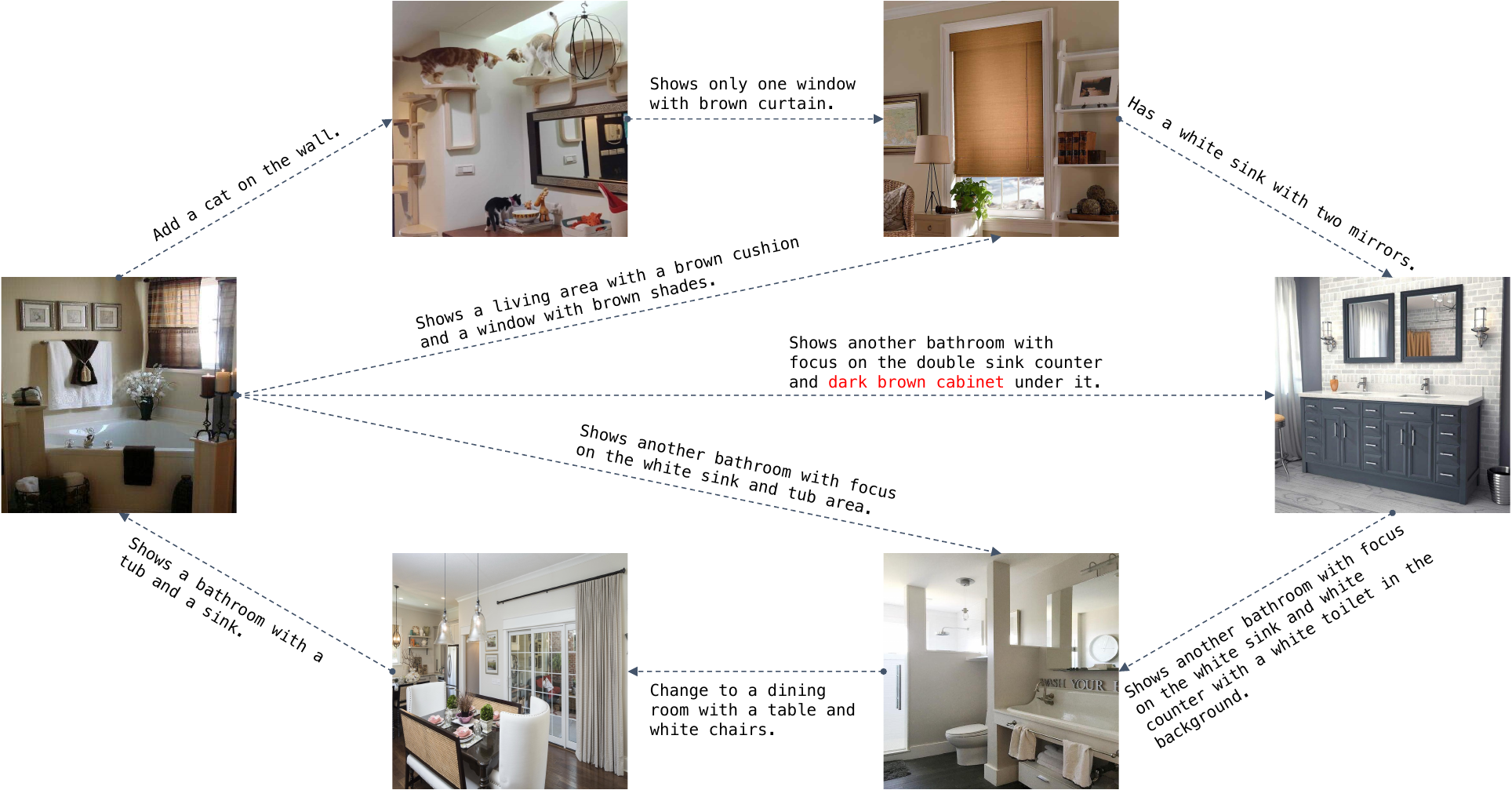}
\caption{Visual delta generation results with VDG on CIRR validation set.}
\label{fig:supp_cirr_val_gen}
\end{figure*}

\begin{figure*}[!t]
\centering
\includegraphics[width=0.99\linewidth]{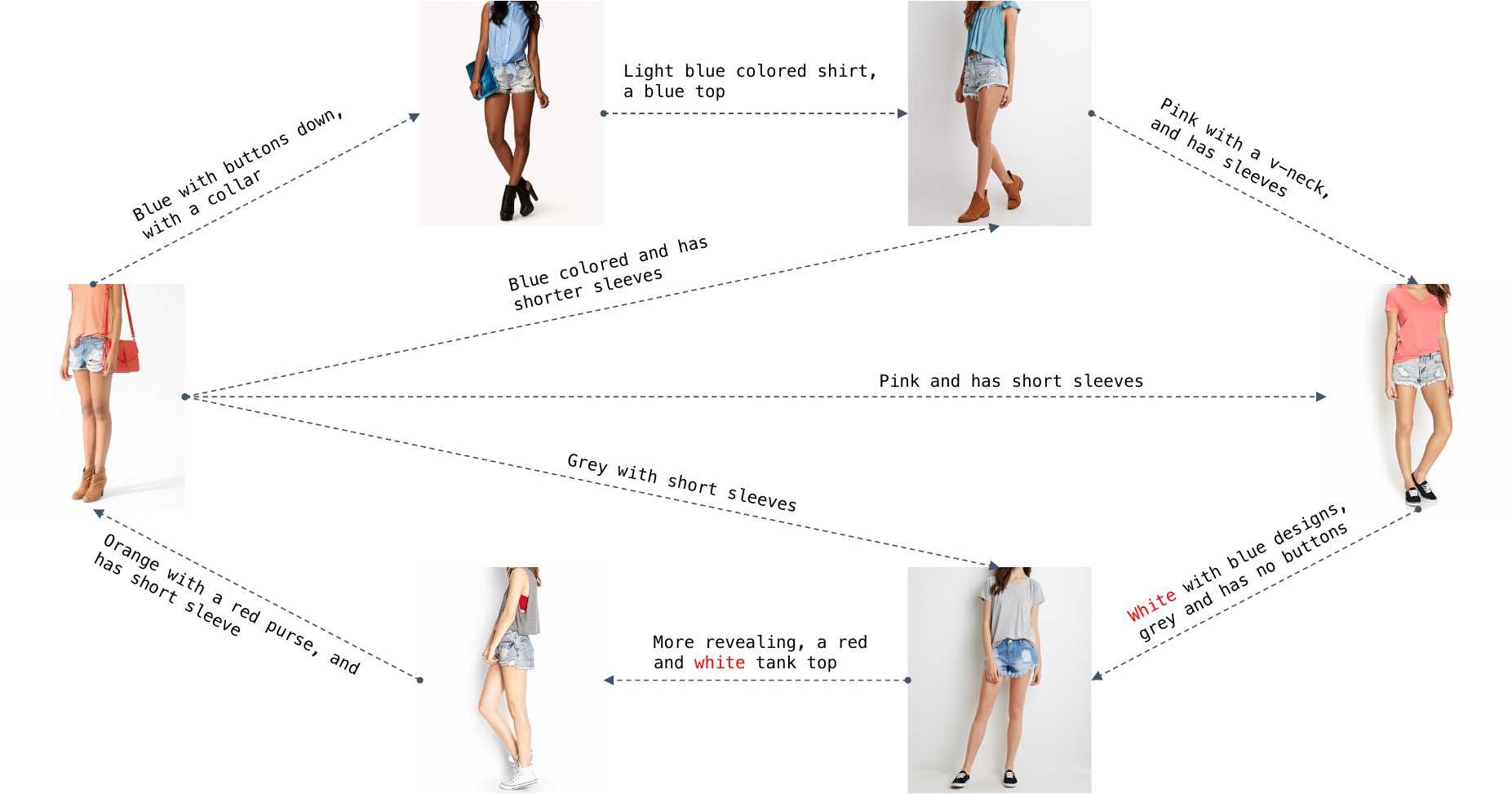}
\caption{Visual delta generation results with VDG on the DeepFashion dataset.}
\label{fig:supp_deepfashion_gen}
\end{figure*}

\paragraph{VDG Generation Results.} We provide more generation results based on subgroups in Figs. \ref{fig:supp_cirr_val_gen} and \ref{fig:supp_deepfashion_gen}. We notice that the VDG excels in generating high-quality visual deltas, with only a few errors.

\begin{figure*}[!t]
\centering
\includegraphics[width=0.99\linewidth]{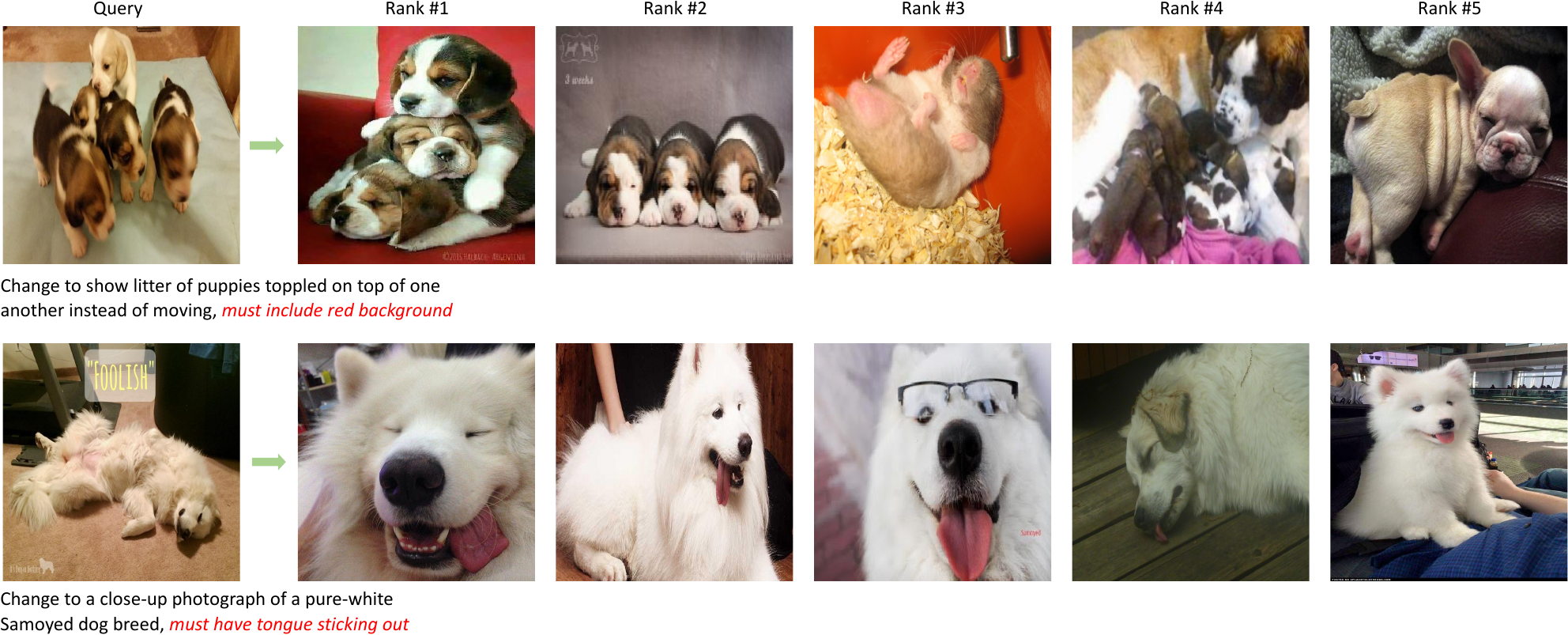}
\caption{Retrieval results on the CIRR test set.}
\label{fig:supp_cirr_test}
\end{figure*}

\begin{figure*}[!t]
\centering
\includegraphics[width=0.99\linewidth]{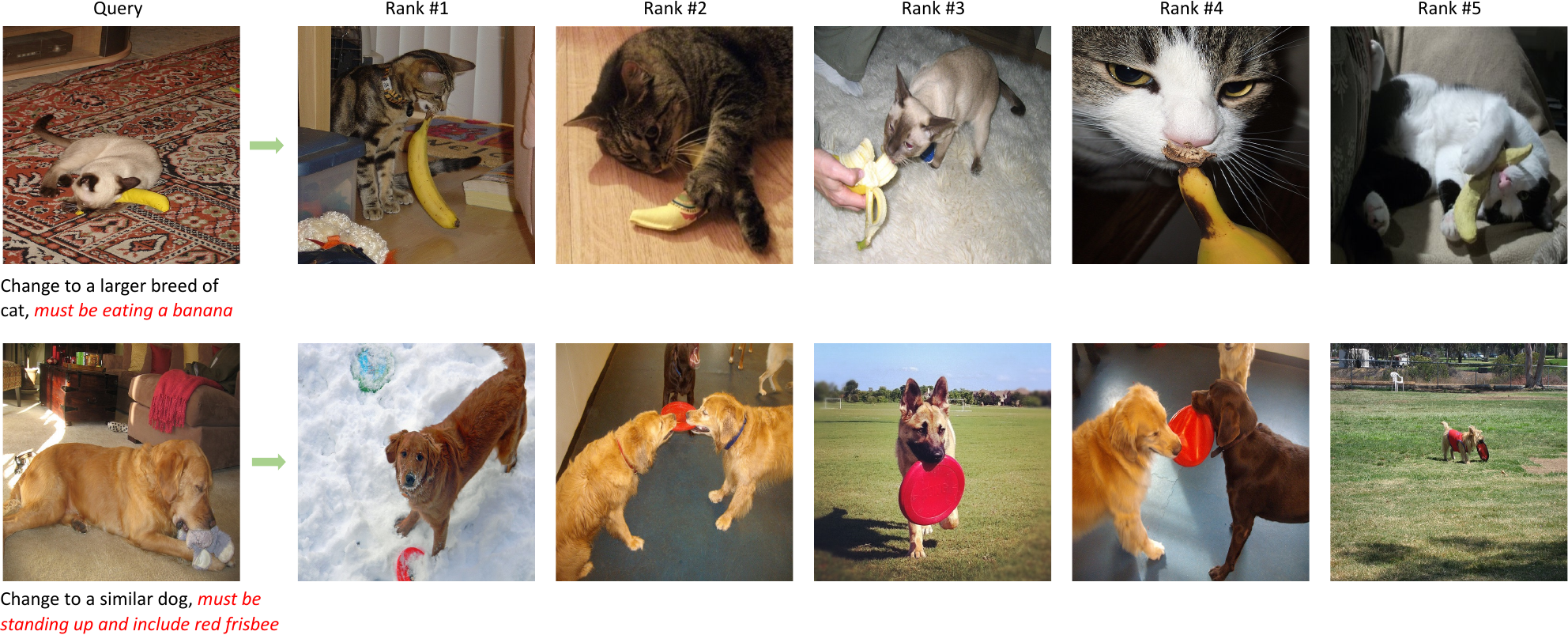}
\caption{Retrieval results on the COCO dataset.}
\label{fig:supp_coco}
\end{figure*}

\begin{figure*}[!t]
\centering
\includegraphics[width=0.99\linewidth]{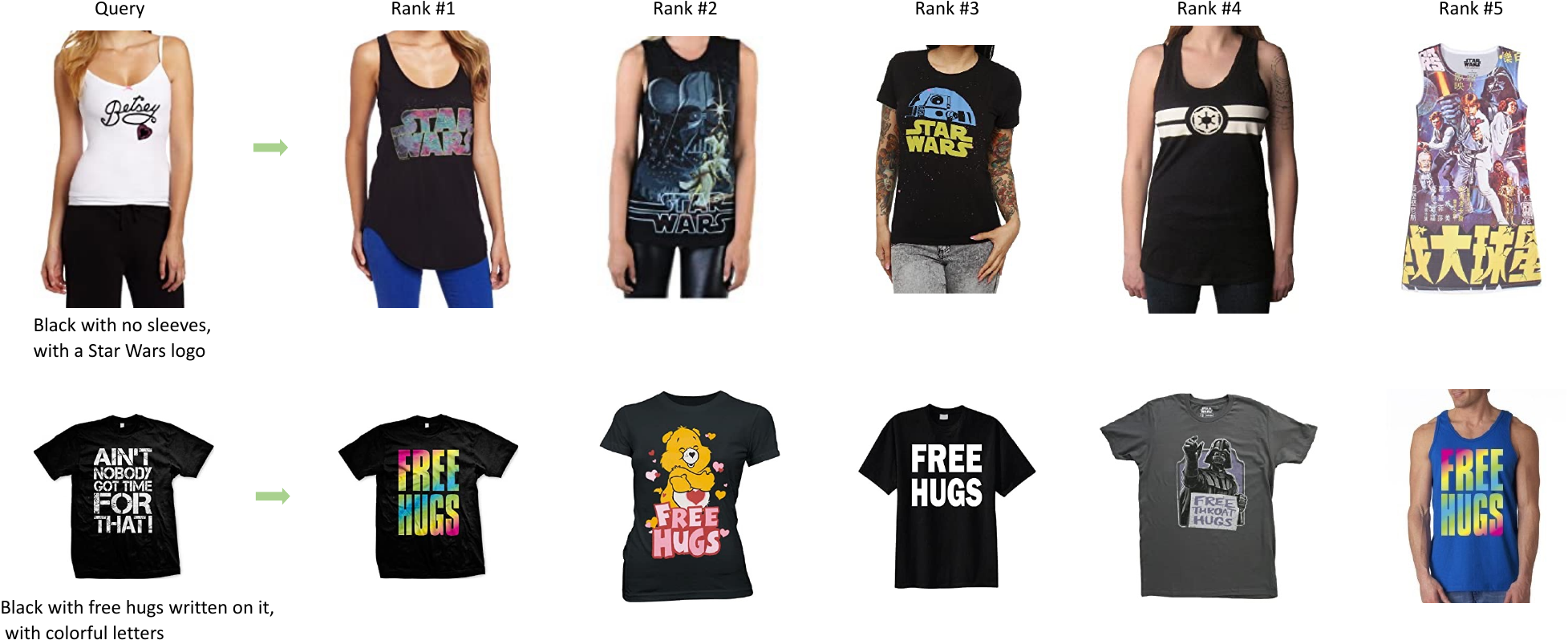}
\caption{Retrieval results on the FashionIQ dataset.}
\label{fig:supp_fashionIQ}
\end{figure*}

\begin{figure*}[!t]
\centering
\includegraphics[width=0.99\linewidth]{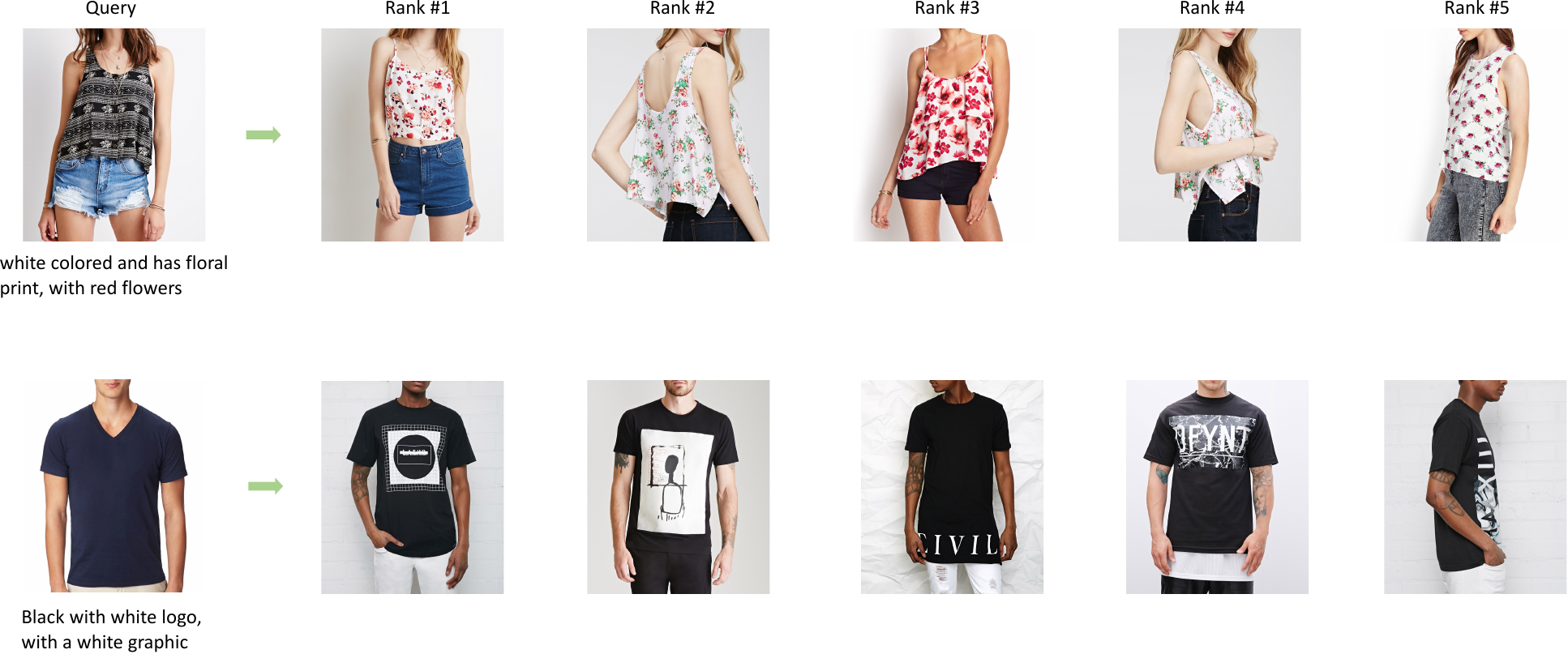}
\caption{Retrieval results on the DeepFashion dataset.}
\label{fig:supp_deepfashion}
\end{figure*}

\paragraph{Retrieval Results.} We provide more retrieval results for natural images in Figs. \ref{fig:supp_cirr_test} and \ref{fig:supp_coco}, and for fashion images in Figs. \ref{fig:supp_fashionIQ} and \ref{fig:supp_deepfashion}. In the domain of natural images, we chose query examples containing the word \textit{must}. We observe that our CIR model effectively grasps the meaning of the text query and reflects this understanding in the retrieval results. In addition, the domain of fashion images is also well-represented in the retrieval results, accurately reflecting the user's text query while maintaining visual information of query image.

